\documentclass[sigconf, nonacm]{acmart}
\usepackage{bbding}
\usepackage{subcaption}
\usepackage{makecell}
\usepackage{multirow}
\usepackage{diagbox} 
\usepackage{enumerate} 
\usepackage{amsmath}
\usepackage{autobreak}
\usepackage{color}
\usepackage[capitalize]{cleveref}
\newcommand{\aref}[2]{\hyperref[#1]{\ref*{#1}{#2}}}
\usepackage{booktabs}

\usepackage[shortlabels]{enumitem}
\usepackage{fontawesome}
\usepackage{array}
\usepackage{xspace}

\newcommand\vldbdoi{10.14778/3648160.3648167}
\newcommand\vldbpages{1241 - 1254}
\newcommand\vldbvolume{17}
\newcommand\vldbissue{6}
\newcommand\vldbyear{2024}
\newcommand\vldbauthors{\authors}
\newcommand\vldbtitle{\shorttitle} 
\newcommand\vldbavailabilityurl{https://github.com/iDC-NEU/NeutronBench}

\newcommand\vldbpagestyle{empty}

\newcommand{\eat}[1]{}
\newcommand{\remove}[1]{}
\newcommand{\reidx}[1]{}

\newcommand{\Paragraph} [1] {\smallskip\noindent{\bf #1. }}
\newcommand{\System}{NeutronBench\xspace}

\newcommand{\mixFanoutSamplerateSpeedup}{1.74\xspace}

\newcommand{\Metisextended}{Metis-extend\xspace}

\newcommand{\mixBatchSizeReddit}{1.64\xspace}
\newcommand{\mixBatchSizeProducts}{1.52\xspace}

\newcommand{\zeroSpeedup}{1.74\xspace}
\newcommand{\pipelineSpeedup}{1.30\xspace}

\newcommand{\ZPSpeedup}{2.26\xspace}

\begin{document}
\title{Comprehensive Evaluation of GNN Training Systems: A Data Management Perspective}

\vspace{-0.25in}


\vspace{-0.35in}
\eat{
\settopmatter{authorsperrow=4}
\author{Hao Yuan}
\affiliation{
  \institution{Northeastern University\country{China} 
  \\ yuanhao@stumail.neu.edu.cn
  }
}

\author{Yajiong Liu}
\affiliation{\institution{Northeastern University\country{China}
\\ liuyajiong@stumail.neu.edu.cn
}}

\author{Yanfeng Zhang}
\affiliation{\institution{Northeastern University\country{China}
\\ zhangyf@mail.neu.edu.cn
}}

\author{Xin Ai}
\affiliation{\institution{Northeastern University\country{China}
\\ aixin0@stumail.neu.edu.cn
}}

\author{Qiange Wang}
\affiliation{\institution{National University of Singapore\country{Singapore}
\\ wang.qg@nus.edu.sg
}}

\author{Chaoyi Chen}
\affiliation{\institution{Northeastern University\country{China}
\\ chenchaoy@stumail.neu.edu.cn
}}

\author{Yu Gu}
\affiliation{\institution{Northeastern University\country{China}
\\ guyu@mail.neu.edu.cn
}}

\author{Ge Yu}
\affiliation{\institution{Northeastern University\country{China}
\\yuge@mail.neu.edu.cn
}}
}

\settopmatter{authorsperrow=4}

\author{Hao Yuan}
\affiliation{
  \institution{Northeastern University\country{China} 
  }
}
\email{yuanhao@stumail.neu.edu.cn}

\author{Yajiong Liu}
\affiliation{\institution{Northeastern University\country{China}}}
\email{liuyajiong@stumail.neu.edu.cn}

\author{Yanfeng Zhang}
\affiliation{\institution{Northeastern University\country{China}}}
\email{zhangyf@mail.neu.edu.cn}

\author{Xin Ai}
\affiliation{\institution{Northeastern University\country{China}}}
\email{aixin0@stumail.neu.edu.cn}

\author{Qiange Wang}
\affiliation{\institution{National University of Singapore\country{Singapore}}}
\email{wang.qg@nus.edu.sg}

\author{Chaoyi Chen}
\affiliation{\institution{Northeastern University\country{China}}}
\email{chenchaoy@stumail.neu.edu.cn}

\author{Yu Gu}
\affiliation{\institution{Northeastern University\country{China}}}
\email{guyu@mail.neu.edu.cn}

\author{Ge Yu}
\affiliation{\institution{Northeastern University\country{China}}}
\email{yuge@mail.neu.edu.cn}

\vspace{-0.15in}






\begin{abstract}

Many Graph Neural Network (GNN) training systems have emerged recently to support efficient GNN training. Since GNNs embody complex data dependencies between training samples, the training of GNNs should address distinct challenges different from DNN training in data management, such as data partitioning, batch preparation for mini-batch training, and data transferring between CPUs and GPUs. These factors, which take up a large proportion of training time, make data management in GNN training more significant.
This paper reviews GNN training from a data management perspective and provides a comprehensive analysis and evaluation of the representative approaches. We conduct extensive experiments on various benchmark datasets and show many interesting and valuable results. We also provide some practical tips learned from these experiments, which are helpful for designing GNN training systems in the future.

\end{abstract}


\maketitle

\vspace{-.1in}
\pagestyle{\vldbpagestyle}
\begingroup\small\noindent\raggedright\textbf{PVLDB Reference Format:}\\
\vldbauthors. \vldbtitle. PVLDB, \vldbvolume(\vldbissue): \vldbpages, \vldbyear.\\
\href{https://doi.org/\vldbdoi}{doi:\vldbdoi}
\endgroup

\begingroup
\renewcommand\thefootnote{}\footnote{\noindent
This work is licensed under the Creative Commons BY-NC-ND 4.0 International License. Visit \url{https://creativecommons.org/licenses/by-nc-nd/4.0/} to view a copy of this license. For any use beyond those covered by this license, obtain permission by emailing \href{mailto:info@vldb.org}{info@vldb.org}. Copyright is held by the owner/author(s). Publication rights licensed to the VLDB Endowment. \\
\raggedright Proceedings of the VLDB Endowment, Vol. \vldbvolume, No. \vldbissue\ %
ISSN 2150-8097. \\
\href{https://doi.org/\vldbdoi}{doi:\vldbdoi} \\
}\addtocounter{footnote}{-1}\endgroup

\vspace{-.15in}
\ifdefempty{\vldbavailabilityurl}{}{
\vspace{.3cm}
\begingroup\small\noindent\raggedright\textbf{PVLDB Artifact Availability:}\\The source code, data, and/or other artifacts have been made available at \url{\vldbavailabilityurl}.
\endgroup
}


\section{Introduction}

Graph Neural Networks (GNNs) are a class of Deep Neural Networks (DNNs) that can effectively process and analyze graph-structured data \cite{GNN_survey}, and it is widely used in a variety of graph-related tasks \cite{GraphSage_NIPS17, GAT_2018, link_2018, link_2018_nips, graph_2018_nips}.
With the increasing size of real-world graph data, it becomes difficult to train a large-scale GNN in a limited memory space. Distributed sample-based mini-batch GNN training emerges as a promising solution \cite{DistDGL_IAAA20, DistDglv2-kdd22, Aligraph_VLDB19, AGL_VLDB20, PaGraph_SoCC20, P3_OSDI21, pytorch_directed_VLDB21, GNNLab_EuroSys22, ByteGNN_VLDB22, SALIENTPlus_MLSYS23,Legion_ATC23,BGL_NIDS23}. In this training approach, the distributed environment offers ample computational and storage resources for training. The sampling reduces the size of the training graph \cite{GraphSage_NIPS17, FastGCN_ICLR18, GraphSAINT_ICLR20}, and the mini-batch training increases the frequency of model update to accelerate model convergence \cite{DistDglv2-kdd22}.

In a multi-node CPU and GPU heterogeneous training scenario, the training process of GNN and DNN can be divided into the following four steps.
(1) Data partitioning: The input graph data is partitioned into multiple parts for distributed training.
(2) Batch preparation: The data assigned to each worker is divided into multiple batches.
(3) Data transferring: The prepared batch data is transferred to the GPU.
(4) NN computation: NN computation is performed on the GPU side, including forward and backward propagation and parameter update.

\begin{figure}
  \centering
  \includegraphics[width=2.6in]{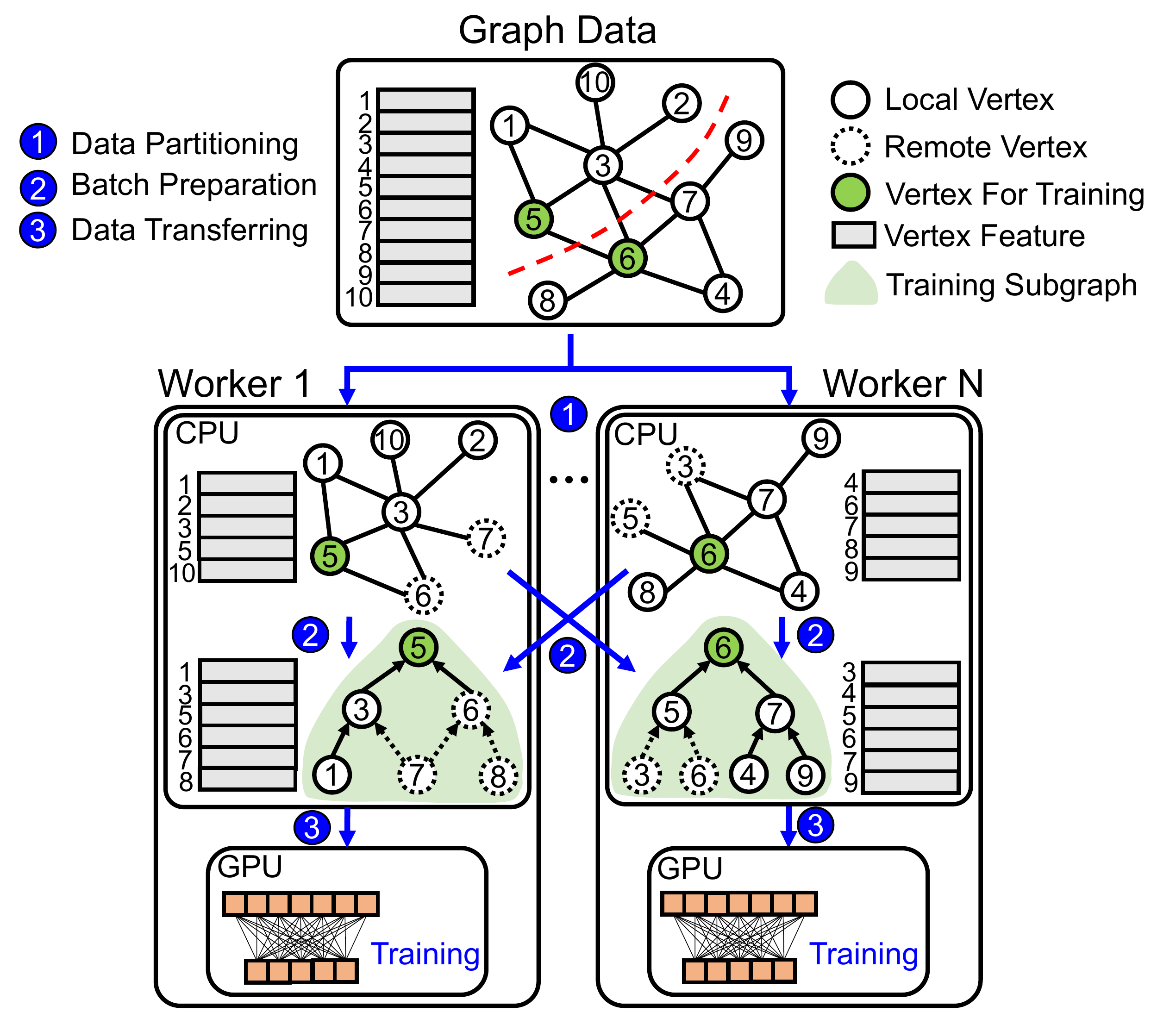}
  \vspace{-.1in}
  \caption{GNN training process.}
  \label{fig:gnn-training-data-perspective}
\vspace{-0.2in}
\end{figure}

Compared with DNNs, GNNs embody complex data dependencies between training samples. These data dependencies pose key challenges to the management of GNN data. Figure \ref{fig:gnn-training-data-perspective} depicts the end-to-end training process of GNNs from the data management perspective.
Due to the dependence between data samples, GNN needs to frequently exchange data, which makes parallel GNN training a bottleneck.
In the data partitioning step, since the data of the DNN is independent, it is only required to divide the data evenly into multiple partitions.
However, for GNN training, the vertices need to access their multi-hop neighbor vertices, and the distribution of these neighbor vertices directly affects GNN training performance. 
In the batch preparation step, DNN only randomly divides the data samples into batches because of no dependency between data samples. GNN generates the training subgraphs by sampling for each batch. Complex data dependencies between data samples increase the sampling overhead.
In the data transferring step, the low PCIe bandwidth between CPU and GPU results in high transfer overheads for both GNN and DNN.
Compared to DNN, there are a large number of duplicate vertices and edges across different batches \cite{PaGraph_SoCC20} due to the complex dependencies between vertices. Transferring such redundant vertices and edges wastes bandwidth resources severely \cite{PaGraph_SoCC20, GNNLab_EuroSys22, SALIENTPlus_MLSYS23}, bringing opportunities for data transfer optimizations.

\begin{figure}
\vspace{-0.1in}
  \centering
  \includegraphics[width=2.5in]{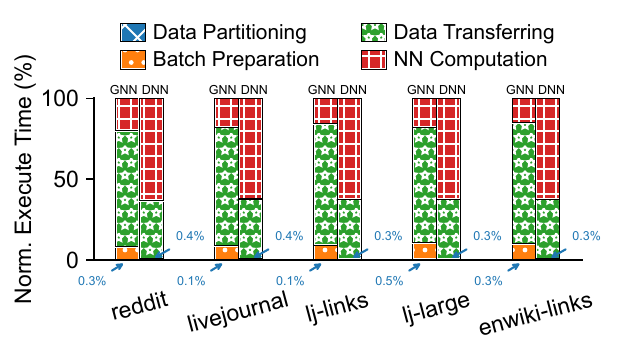}
  \vspace{-0.18in}
  \caption{Time portion of different steps in GNN training and DNN training.}
  \label{fig_exp_performance_breakdown}
\vspace{-0.18in}
\end{figure}

To illustrate the impact of handling data dependencies, Figure \ref{fig_exp_performance_breakdown} shows a step-level time breakdown in both GNN and DNN training. For GNN training, we use a two-layer GCN \cite{GCN_ICLR17} with a two-layer multi-layer perceptron model (MLP). 
For DNN training, to ensure fairness, we use a two-layer MLP with the same parameter settings as that in GCN. 
Since data partitioning is a preprocessing task performed only once before training, its runtime is ignorable.
We can observe that the NN computation step consumes the majority of the runtime in DNN training, while the training in GNN only takes up a small portion. 
Furthermore, the overhead of the data management steps in GNN training (data partitioning, batch preparation, and data transferring) is notably higher than that in DNN training.
Compared to DNN training, GNN training involves processing large-scale graph data. The data samples in GNN (i.e., graph vertices) exhibit complex dependencies, making the data management steps of GNN training more intricate and time-consuming than those of DNN training. Therefore, efficient data management is very important for GNN training.

Many GNN training systems have recently emerged to support efficient GNN training \cite{DistDGL_IAAA20, PyG_ICLR19, ROC_MLSys20, P3_OSDI21, Dorylus_OSDI21, PaGraph_SoCC20, GNNLab_EuroSys22, ByteGNN_VLDB22, NeutronStar_SIGMOD22, hongtu_SIGMOD_2024, MariusGNN_EuroSys23, Legion_ATC23, BGL_NIDS23, G3_SIGMOD_2023, Dsp_ppopp_2023, Ducati_sigmod_2023, Betty_ASPLOS23, NeutronStream_VLDB23}. 
These systems differ greatly in their targeted application scenarios and optimization techniques, especially from a data management perspective. 
The importance of data management in GNN training and the recent emergence of various data management techniques motivate us to study the impact of different optimizations and parameters in GNN training. In this paper, we review the training process of GNN from the data management perspective and provide a comprehensive analysis and evaluation of optimization techniques proposed in GNN training systems. The contributions of this paper can be summarized as follows.
\begin{itemize}[leftmargin=*]
    \item A taxonomy of data management techniques in GNN training. 
    \item A comprehensive evaluation of data management techniques in GNN training.
    \item A summarization of lessons learned from our evaluation results.
\end{itemize}
We believe that our comprehensive analysis and evaluation results should be helpful for researchers and system developers to further improve the existing GNN training systems or design new GNN training systems.

\newcommand{\gnnPartitionT}{0.27\%\xspace} 
\newcommand{\gnnBatchT}{34.33\%\xspace}
\newcommand{\gnnTransT}{46.42\%\xspace}
\newcommand{\gnnNNT}{18.98\%\xspace}

\newcommand{\gnnPartition}{0.24\%\xspace} 
\newcommand{\gnnBatch}{9.09\%\xspace}
\newcommand{\gnnTrans}{73.39\%\xspace}
\newcommand{\gnnNN}{17.27\%\xspace}

\newcommand{\dnnPartition}{0.30\%\xspace}
\newcommand{\dnnBatch}{0.32\%\xspace}
\newcommand{\dnnTrans}{36.90\%\xspace}
\newcommand{\dnnNN}{62.48\%\xspace}

\newcommand{\gnnPartitionTime}{139.11s\xspace}
\newcommand{\dnnPartitionTime}{4.62s\xspace}
\newcommand{\motivationBatchSize}{2048\xspace}

\Paragraph{Related Work}
Recent surveys \cite{GNN_SYS_Survey, GNN_SYS_Servey2, GNNDNNEvolution_Arxiv23} provide an overview of the state-of-the-art GNN training systems and related optimization techniques.
Unlike these survey papers, we categorize and summarize various optimizations for GNN training from a data management perspective, and provide a comprehensive experimental evaluation of them.
Huang et al. investigate the performance gap of GPU-based GNN systems and provide a set of optimizations to enhance performance \cite{GNNComputing_PPoPP21}. A recent work \cite{GNNEvalPartition_VLDB23} empirically studies graph partitioning for distributed GNN training. These two works only evaluate a single step of the entire GNN training process and do not separate data management from the entire training process for evaluation.
Liu et al. provide a thorough experimental comparison of different sampling algorithms \cite{Sample_survey_CAA22}. Wang et al. experimentally evaluate GNN training and inference on a single GPU \cite{GNNEval_Neurocomputing21}. These two works focus on single GPU GNN training and do not consider a distributed environment.

\section{GNN Training Process}
\label{sec_background}

GNNs are a type of neural network that operates on graphs. The key idea behind GNNs is to learn vertex and edge representations that capture the structural information of the graph. 

\Paragraph{Graph Neural Network (GNN)} Similar to traditional neural networks, the training process of GNNs includes forward propagation and backward propagation. In forward propagation, each vertex collects its neighbors' features to generate the aggregation result using an aggregation function:
\begin{align}
    a_{v}^{(l)} &=AGGREGATE(\{h_u^{(l-1)} | {u} \in {N(v)}\}),
\end{align}
\noindent where $h_u^{(l-1)}$ denotes the embedding of vertex $u$ at ($l$-1)-th layer, $N(v)$ denotes the incoming neighbors of vertex $v$, and $a_v^{(l)}$ denotes the aggregation result of vertex $v$ at $l$-th layer. Specifically, $h_v^0$ denotes the input feature of vertex $v$. The aggregate functions can be \texttt{sum}, \texttt{average}, \texttt{max/min}, etc. Subsequently, the aggregated features are combined with the vertex's own features using a combine function:
\begin{align}
    h_v^{l}&=\sigma(\mathbf{W}^{(l)} \cdot COMBINE(h_v^{l-1}, a_{v}^l)),
\end{align}
\noindent where $\sigma$ denotes a non-linear function (e.g., \texttt{ReLU}), $\mathbf{W}^{(l)}$ denotes the weight matrix that transforms the vertex embedding at ($l$-1)-th layer. The \texttt{combine} function can be concatenation, element-wise multiplication, or summation. 
Then, a neural network (usually a multi-layer perceptron model, MLP) is used to update the features of each vertex. Finally, the output vertex features are compared to the ground truth labels to compute the loss.
In the backward propagation, the loss is propagated through the neural network in the reverse direction, generating gradients that are used to update the model's parameters.

\Paragraph{Distributed Mini-Batch GNN Training}
Due to the increasing size of real-world graph data, many emerging GNN training systems \cite{DistDGL_IAAA20, DistDglv2-kdd22, Aligraph_VLDB19, AGL_VLDB20, PaGraph_SoCC20, P3_OSDI21, pytorch_directed_VLDB21, GNNLab_EuroSys22, ByteGNN_VLDB22, SALIENTPlus_MLSYS23, Legion_ATC23, BGL_NIDS23} adopt distributed mini-batch training. This approach splits the training vertices into multiple mini-batches and conducts GNN training among workers. 
Figure \ref{fig:gnn-training-data-perspective} illustrates an example of the distributed mini-batch training for training a 2-layer GNN on vertices $V_5$ and $V_6$. In the data partitioning step, the graph data are stored in a distributed graph storage, and each worker takes a subset of vertex samples for training. In the batch preparation step, each worker samples the $L$-hop neighbors of training vertices to generate \textit{training subgraphs}. For example, as shown in Figure \ref{fig:gnn-training-data-perspective}, the training subgraph of $V_5$ is sampled 2-hop subgraphs rooted from vertex $V_5$ (green shaded).
The key idea of this step is to let each worker make the remote-dependent neighbors readily prepared locally before training starts. As a $L$-layer GNN, this approach needs to retrieve not only the vertex's direct in-neighbors but also its $\{2, ..., k\}$-hop in-neighbors. 
For example, $V_6$, $V_7$, and $V_8$ are remote-dependent neighbors of $V_5$, and worker 1 needs to fetch their feature vectors before training starts. Note that the sampled vertices may be deduplicated ($V_7$ in worker 1 is simultaneously sampled by $V_3$ and $V_6$). 
In the data transferring step, the training data, including sampled subgraphs and corresponding feature vectors, are loaded from the CPU to the local GPU through a PCIe interconnect.
In the NN computation step, with the $L$-hop sampled subgraph locally, this approach performs normal forward/backward propagation layer-by-layer within a worker without any communication. Only the gradients need to be synchronized in the parameter update stage.

\begin{table*}[t]
\vspace{-0.3in}
\footnotesize
\caption{Summary of representative GNN training system and data management techniques.}
\vspace{-0.1in}
\label{tab:summary_gnn_system}
\begin{tabular}{c|c|c|c|c|c|c|c|c|c}
\hline
\multirow{2}*{\textbf{Year}} & \multirow{2}*{\textbf{Systems}}& \multirow{2}*{\makecell{\textbf{Deploy Platform}}} & \makecell{\textbf{data partitioning}} & \multicolumn{3}{c|}{\textbf{Batch Preparation}} & \multicolumn{3}{c}{\textbf{Data Transferring}} \\ 
\cline{4-10}
& & & \textbf{Partition Method} & \textbf{Train Method} & \textbf{Sample} & \textbf{Sample Method} & \textbf{Transfer Method} & \textbf{Pipeline} & \textbf{Cache}\\
\hline

2019 & DGL \cite{DGL_Arxiv19} & Multi-GPU & N/A & Mini-batch & \CheckmarkBold & Fanout-based & Extract-Load & \CheckmarkBold & \XSolidBrush \\ \hline
2019 & PyG \cite{PyG_ICLR19} & Multi-GPU & N/A & Mini-batch & \CheckmarkBold & Fanout-based & Extract-Load & \XSolidBrush & \XSolidBrush \\ \hline
2019 & AliGraph \cite{Aligraph_VLDB19} & CPU-cluster & \makecell{Hash/Metis\\/Streaming} & Mini-batch & \CheckmarkBold & \makecell{Fanout-based/\\Ratio-based} & N/A & \XSolidBrush & \XSolidBrush \\ \hline
2019 & NeuGraph \cite{Neugraph_ATC19} & Multi-GPU & Hash & Full-batch & \XSolidBrush & N/A & Extract-Load & \XSolidBrush & \XSolidBrush \\ \hline
2020 & AGL \cite{AGL_VLDB20} & CPU-cluster & Hash & Mini-batch & \CheckmarkBold & Fanout-based & N/A & \XSolidBrush & \XSolidBrush \\ \hline
2020 & DistDGL \cite{DistDGL_IAAA20} & CPU-cluster & \Metisextended & Mini-batch & \CheckmarkBold & \makecell{Fanout-based/\\Ratio-based} & N/A & \CheckmarkBold & \XSolidBrush \\ \hline
2020 & ROC \cite{ROC_MLSys20} & GPU-cluster & Hash & Full-batch & \XSolidBrush & N/A & Extract-Load & \XSolidBrush & \XSolidBrush \\ \hline
2020 & PaGraph \cite{PaGraph_SoCC20} & Multi-GPU & Streaming & Mini-batch & \CheckmarkBold & Fanout-based & Extract-Load & \XSolidBrush & \CheckmarkBold \\ \hline
2021 & P3 \cite{P3_OSDI21} & GPU-cluster & Hash & Mini-batch & \CheckmarkBold & Fanout-based & Extract-Load & \XSolidBrush & \XSolidBrush \\ \hline
2021 & DistGNN \cite{DistGNN_SC21} & CPU-cluster & Hash & Full-batch & \XSolidBrush & N/A & N/A & \XSolidBrush & \XSolidBrush \\ \hline
2021 & DGCL \cite{DGCL_EuroSys21} & GPU-cluster & Hash & Full-batch & \XSolidBrush & N/A & Extract-Load & \XSolidBrush & \XSolidBrush \\ \hline
2021 & Dorylus \cite{Dorylus_OSDI21} & Serverless & Hash & Full-batch & \XSolidBrush & N/A & N/A & \CheckmarkBold & \XSolidBrush \\ \hline
2021 & Pytorch-direct \cite{pytorch_directed_VLDB21} & Multi-GPU & N/A & Mini-batch & \CheckmarkBold & Fanout-based & GPU direct access & \CheckmarkBold & \XSolidBrush \\ \hline
2022 & GNNLab \cite{GNNLab_EuroSys22} & Multi-GPU & N/A & Mini-batch & \CheckmarkBold & Fanout-based & Extract-Load & \CheckmarkBold & \CheckmarkBold \\ \hline
2022 & ByteGNN \cite{ByteGNN_VLDB22} & CPU-cluster & Streaming & Mini-batch & \CheckmarkBold & Fanout-based & N/A & \CheckmarkBold & \XSolidBrush \\ \hline
2022 & BNS-GCN \cite{BNS-GCN-MLSYS22} & GPU-cluster & Metis & Full-batch & \CheckmarkBold & Ratio-based & Extract-Load & \XSolidBrush & \XSolidBrush \\ \hline
2022 & DistDGLv2 \cite{DistDglv2-kdd22} & GPU-cluster & \Metisextended & Mini-batch & \CheckmarkBold & Fanout-based & Extract-Load & \CheckmarkBold & \XSolidBrush \\ \hline
2022 & NeutronStar \cite{NeutronStar_SIGMOD22} & GPU-cluster & Hash & Full-batch & \XSolidBrush & N/A & Extract-Load & \XSolidBrush & \XSolidBrush \\ \hline
2022 & Sancus \cite{Sancus_VLDB22} & GPU-cluster & Hash & Full-batch & \XSolidBrush & N/A & Extract-Load & \XSolidBrush & \CheckmarkBold \\ \hline
2022 & SALIENT \cite{SALIENT_MLSYS22} & Multi-GPU & N/A & Mini-batch & \CheckmarkBold & Fanout-based & GPU direct access & \CheckmarkBold & \XSolidBrush \\ \hline
2023 & Betty \cite{Betty_ASPLOS23} & GPU-only & Metis & Mini-batch & \CheckmarkBold  & Fanout-based & Extract-Load & \XSolidBrush & \XSolidBrush \\ \hline
2023 & MariusGNN \cite{MariusGNN_EuroSys23} & GPU-only & Hash & Mini-batch & \CheckmarkBold & Fanout-based & Extract-Load & \CheckmarkBold & \XSolidBrush \\ \hline
2023 & Legion \cite{Legion_ATC23} & Multi-GPU & Metis/Hash & Mini-batch & \CheckmarkBold & Fanout-based & Extract-Load & \CheckmarkBold & \CheckmarkBold \\ \hline
2023 & SALIENT++ \cite{SALIENTPlus_MLSYS23} & GPU-cluster & \Metisextended & Mini-batch & \CheckmarkBold & Fanout-based & GPU direct access & \CheckmarkBold & \CheckmarkBold \\ \hline
2023 & BGL \cite{BGL_NIDS23} & Multi-GPU & Streaming & Mini-batch & \CheckmarkBold & Fanout-based & Extract-Load & \CheckmarkBold & \CheckmarkBold \\ \hline

\end{tabular}
\vspace{-0.1in}
\end{table*}

\section{Taxonomy of Data Management Techniques in GNN Training}
\label{sec:taxonomy}

There has been a surge of GNN training systems emerging in academia and industry to support efficient training. A variety of novel data management techniques have been proposed to accelerate the training of GNNs. This section provides a taxonomy of these techniques. As shown in Table \ref{tab:summary_gnn_system}, the recently emerged systems are listed from four aspects, including deployment platform, data partitioning, batch preparation, and data transferring.

\Paragraph{Deployment Platform} 
From the perspective of deployment platforms, CPU-cluster \cite{Aligraph_VLDB19,AGL_VLDB20,DistDGL_IAAA20,DistGNN_SC21,ByteGNN_VLDB22}, Multi-GPU \cite{DGL_Arxiv19,PyG_ICLR19,Neugraph_ATC19,PaGraph_SoCC20,GNNLab_EuroSys22,pytorch_directed_VLDB21,SALIENT_MLSYS22,Legion_ATC23,BGL_NIDS23}, and GPU-cluster \cite{ROC_MLSys20,P3_OSDI21,DGCL_EuroSys21,BNS-GCN-MLSYS22,DistDglv2-kdd22,NeutronStar_SIGMOD22,Sancus_VLDB22,SALIENTPlus_MLSYS23} are the most commonly used. CPU-cluster is a network of multiple computers in which the CPU is the only computing component. Multi-GPU refers to the configuration of multiple GPUs inside a single compute node. GPU-cluster is similar to a CPU-cluster, but each compute node is equipped with one or more GPUs.

\Paragraph{Data Partitioning}
In terms of data partitioning, we categorize the data partitioning methods of existing GNN training systems into Hash, \Metisextended, and Streaming. 
Hash is a general graph partitioning method. Different mapping rules are used to meet various task requirements, such as hashing by vertices \cite{Aligraph_VLDB19, Euler, AGL_VLDB20, ROC_MLSys20, DGCL_EuroSys21, Dorylus_OSDI21, NeutronStar_SIGMOD22, Legion_ATC23} or hashing by edges \cite{DistGNN_SC21, Sancus_VLDB22, Neugraph_ATC19, MariusGNN_EuroSys23}.
Metis is an iterative data partitioning method that uses community detection and iterative clustering for partitioning to minimize the amount of communication between partitions. Metis has been widely used in GNN training systems \cite{Aligraph_VLDB19, BNS-GCN-MLSYS22, Legion_ATC23}. In addition, some GNN training systems \cite{DistDGL_IAAA20, DistDglv2-kdd22, SALIENTPlus_MLSYS23} combine Metis with sample-based GNN training, and we label these systems as \Metisextended. 
Streaming data partitioning creates graph partitions in a pass over the sequence of edges, so the decision to assign edges to partitions is made on the fly. By defining different score functions, streaming can achieve more fine-grained data partitioning \cite{PaGraph_SoCC20, ByteGNN_VLDB22, BGL_NIDS23}. For those systems that do not perform data partitioning, we mark them as N/A.

\Paragraph{Batch Preparation}
Regarding batch preparation, GNN training systems are categorized in terms of training method, whether they support sampling, and the adopted sampling method.
Batch size denotes the number of vertices involved in training within each batch and also determines the frequency of model parameter updates. In terms of batch size, GNN training methods can be categorized into two types: full-batch \cite{Neugraph_ATC19, ROC_MLSys20, NeutronStar_SIGMOD22, DistGNN_SC21, Dorylus_OSDI21, DGCL_EuroSys21, BNS-GCN-MLSYS22, Sancus_VLDB22, PaSca_WWW22, DBLP:conf/eurosys/WangY0YCYYZ21} and mini-batch \cite{DistDGL_IAAA20, Aligraph_VLDB19, GNNLab_EuroSys22, ByteGNN_VLDB22, SALIENTPlus_MLSYS23, BGL_NIDS23, DBLP:journals/corr/abs-2202-00075, DBLP:conf/ppopp/SongJ22}. With the increase of graph data size in the real world, the full-batch training method suffers from inefficiency and poor scalability \cite{DistDglv2-kdd22, ByteGNN_VLDB22, BGL_NIDS23}. 
The sample-based mini-batch training method can effectively reduce the size of the training graph \cite{GraphSage_NIPS17, FastGCN_ICLR18, GraphSAINT_ICLR20, ClusterGCN_KDD19}, thus becoming the mainstream training approach \cite{ByteGNN_VLDB22}.
We categorize the sampling methods into two types: fanout-based and ratio-based, which are used to determine the size of the sampling subgraph. The fanout-based sampling method samples a fixed number of neighbors \cite{DistDGL_IAAA20, GraphSage_NIPS17}, while the ratio-based sampling method samples neighbors or the whole graph by a ratio \cite{ClusterGCN_KDD19, GraphSAINT_ICLR20, BNS-GCN-MLSYS22, DistDGL_IAAA20}. 
For systems that do not support sampling, we mark it as N/A in the sample method column.

\Paragraph{Data Transferring}
Regarding data transferring, they are classified in terms of transfer method, whether they support pipeline optimization \cite{pytorch_directed_VLDB21, DistDglv2-kdd22, MariusGNN_EuroSys23, Legion_ATC23}, and whether they support GPU cache optimization \cite{PaGraph_SoCC20, GNNLab_EuroSys22, BGL_NIDS23}. The data transfer between CPU and GPU can be summarized into two categories: extract-load and GPU direct access.
In mini-batch training, after the CPU obtains the sampled subgraph through sampling, it first extracts the vertex features from the feature matrix according to the sampled subgraph. It then loads the sampled subgraph and vertex features to the GPU for training. We call this process "Extract-Load" \cite{PaGraph_SoCC20, GNNLab_EuroSys22}.
GPU direct access \cite{pytorch_directed_VLDB21, SALIENTPlus_MLSYS23} is another data transfer method that uses implicit data transfer \cite{zero_copy_doc, unified_memory_blog} to enable the GPU to access the CPU's memory directly, avoiding explicit data transfers and additional feature extraction operations.
Pipelining decomposes a task into multiple stages and allows these stages to be executed in parallel on different processors.
GPU caching is a cache-based data reuse method \cite{PaGraph_SoCC20, Legion_ATC23, SALIENT_MLSYS22, BGL_NIDS23, Sancus_VLDB22} that effectively reduces the data transfer overhead between CPU and GPU by caching vertex features in GPU memory.
For systems that do not support GPU training, we mark it as N/A in the transfer method column.

\section{Experimental Setup}
\label{sec:setup}

\begin{table}[!t]
\caption{Dataset description.}
\vspace{-0.1in}
\label{tab:Dataset}
\centering
 \footnotesize
{\renewcommand{\arraystretch}{1.2}
\begin{tabular}{c c c c c c}
\hline
{\textbf{Dataset}} &
{\textbf{|V|}} &
{\textbf{|E|}}  &
{\textbf{\#F}}&
{\textbf{\#L}}&
{\textbf{\#hidden}} 
\\
\hline
Reddit \cite{GraphSage_NIPS17} & 232.96K & 114.85M & 602 & 41 & 128 \\
OGB-Arxiv \cite{ogb-arxiv} & 169.34K & 2.48M & 128 & 40 & 128  \\
OGB-Products \cite{ogb-products} & 2.45M & 126.17M & 100 & 47 & 128  \\ 
OGB-Papers \cite{ogb-papers} & 111.06M & 1.6B & 128 & 172 & 128 \\
Amazon \cite{GraphSAINT_ICLR20} & 1.57M & 264.34M & 200 & 107 & 128  \\
LiveJournal \cite{liveJournal_paper} & 4.85M & 90.55M & 600 & 60 & 128 \\
Lj-large \cite{lj-large_paper} & 7.49M & 232.1M & 600 & 60 & 128 \\
Lj-links \cite{lj-link_dataset} & 5.2M & 205.25M & 600 & 60 &128 \\
Enwiki-links \cite{enwiki-links_dataset} & 13.59M & 1.37B & 600 & 60 & 128 \\
\hline
\end{tabular}
}
\vspace{-0.25in}
\end{table}


\Paragraph{Environments}
Our experiments are conducted on the Aliyun ECS cluster with 4 GPU nodes. Each node (ecs.gn6i-c40g1.10xlarge) is equipped with 40 vCPU, 155 GiB DRAM, Intel Xeon (Skylake) Platinum 8163, and an NVIDIA Tesla T4 GPU with 68 SMs, 2560 cores, and 16GB GDDR6 global memory. The host side is running Ubuntu 20.04 LTS OS. The network bandwidth is 10 Gbps/s. Libraries CUDA 11.3, OpenMPI-3.3.2, Pytorch~\cite{Paszke_PyTorch_An_Imperative_2019} v1.9 backend, and cuDNN 8.4 are used. All the codes are compiled with O3 optimization.

\Paragraph{GNN Models}
The models used in our experiments are two representative GNN models, Graph Convolutional Network (GCN) \cite{GCN_ICLR17} and GraphSage \cite{GraphSage_NIPS17}. 
There are also some complex GNN models, e.g., MixHop \cite{Mixhop_ICML19}. Unlike traditional GNN models, it directly aggregates multiple-hop neighbors at each layer. Despite differences in their computation methods, these models share a common L-hop neighbor access pattern. Therefore, we only use GCN and GraphSage as the default GNN models for experiments. Following the suggestion from prior works \cite{DGL_Arxiv19, PaGraph_SoCC20, GNNLab_EuroSys22}, the dimension of the hidden layers is set to 128. For all experiments, the default fanout is (25, 10), and the default batch size is 6000 for all experiments.

\Paragraph{Datasets}
As listed in Table \ref{tab:Dataset}, we use nine real-world graph datasets.
Among them, OGB-Arxiv, Amazon, LiveJournal, Lj-large, Lj-links, and Enwiki-links have skewed degree distributions. Reddit, Amazon, and Enwiki-links have average degrees relatively higher than other large real-world graphs. OGB-Arxiv, OGB-Products, and OGB-Paper have low feature dimensions. Reddit and OGB-Products are two dense graphs with a high clustering coefficient.
In the data partitioning and batch preparation experiments, we use datasets with ground-truth labels, including Reddit, OGB-Arxiv, OGB-Products, and Amazon, because we need to study their effect on the model accuracy. 
Regarding the data transferring experiments, we use different types of datasets, including a social network (Reddit), a citation network (OGB-Papers), co-purchasing networks (OGB-Products and Amazon), livejournal communication networks (Livejournal, Lj-large, and Lj-links), and a wikipedia links network (Enwiki-links). These datasets have different characteristics, e.g., average degree, power-law, and feature dimension, which are commonly used by existing GNN training systems in performance evaluation \cite{PaGraph_SoCC20, GNNLab_EuroSys22, pytorch_directed_VLDB21}.

\Paragraph{Evaluation System}
For data partitioning experiments, we use DistDGL \cite{DistDGL_IAAA20}, which is commonly used by existing GNN training systems in performance evaluation experiments, as our evaluation system. 
For batch preparation and data transferring experiments, we use NeutronStar \cite{NeutronStar_SIGMOD22} as our evaluation system. NeutronStar provides a rich programming interface to facilitate the implementation of different performance optimizations. 

\section{Data Partitioning}
\label{sec_data_partition}


\subsection{Goals and Challenges}
\label{sec-partiton-goal-challenge}
\begin{figure}
\vspace{-0.22in}
  \centering
  \includegraphics[width=3.3in]{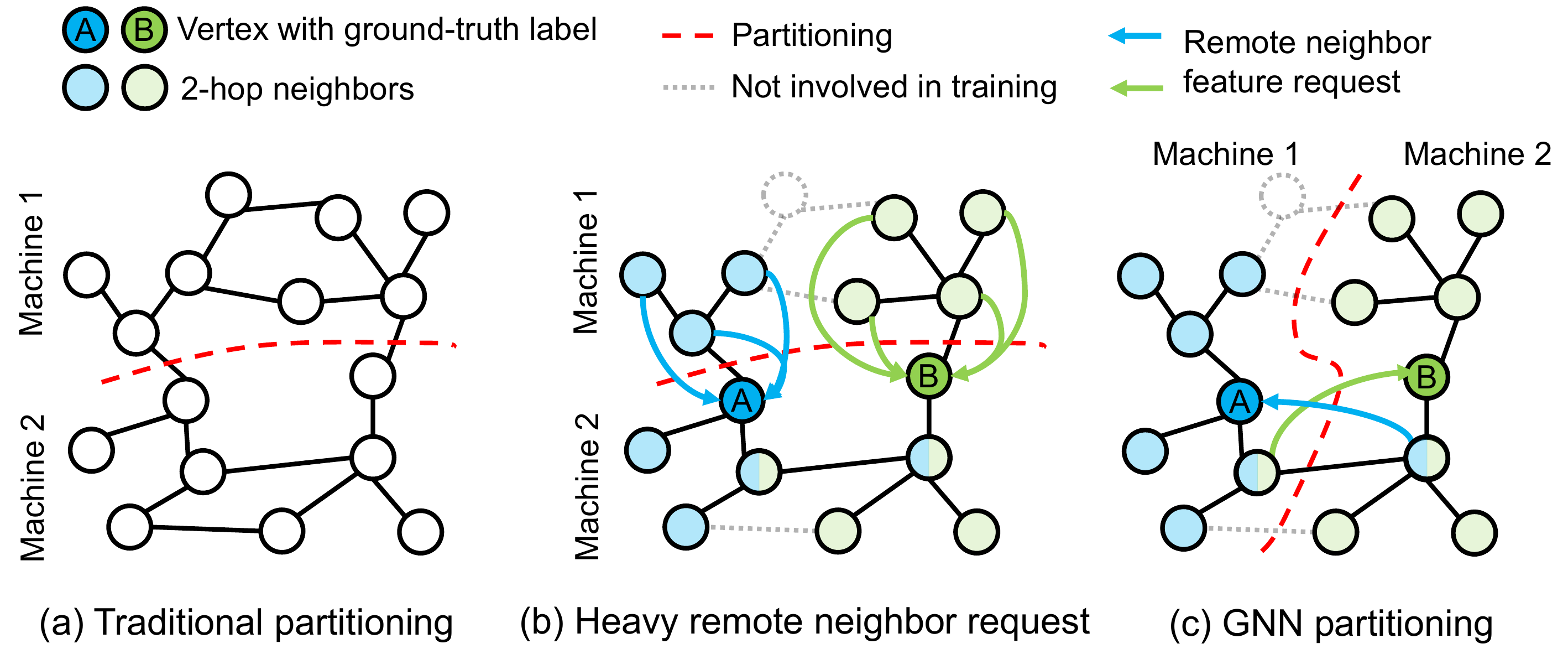}
\vspace{-0.15in}
  \caption{Traditional partitioning vs. GNN partitioning.}
  \label{fig:gnn-partitioning}
\vspace{-0.18in}
\end{figure}

In a distributed setting, the goal of traditional graph partitioning is typically twofold: minimizing communication and balancing the computational load. 
In traditional iterative graph computations, typically \textit{all vertices} and their \textit{1-hop neighbors} (connected through edges) are involved in the computation (e.g., PageRank \cite{PageRank98}).
However, in GNN training, only a portion of vertices with ground-truth labels (i.e., training vertices) and their L-hop subgraphs are involved. For example, in a 2-layer GNN, as shown in Figure \aref{fig:gnn-partitioning}{c}, A and B represent the labeled vertices. The light blue and light green vertices are their 2-hop neighbors. The dashed lines indicate vertices and edges that are not involved in training. 

Due to these differences, traditional graph partitioning methods cannot be directly applied to GNN training. For traditional graph computation, the partitioning in Figure \aref{fig:gnn-partitioning}{a} is optimal because it not only balances the number of vertices and edges but also leads to the minimum number of cutting edges. 
However, this partitioning is not optimal for GNN training due to the following two issues: 
(1) Severe computational load imbalance. The graph partitioner assigns all labeled vertices (A and B) to machine 2, leaving machine 1 idle, thus wasting computational resources. 
(2) High communication volume. The minimum edge-cut graph partitioning method does not consider the distribution of L-hop neighbors of vertices. The neighbors of labeled vertices are scattered in different partitions, which results in heavy remote neighbor requests. 
For example, as shown in Figure \aref{fig:gnn-partitioning}{b}, for the GNN training, using the minimum edge-cut partitioning method from Figure \aref{fig:gnn-partitioning}{a} would result in approximately half of the 2-hop neighbors of training vertices (A and B) being distributed across different partitions. 
This would lead to significant communication overhead, especially considering that the vertices of a GNN have high-dimensional feature representations.
In contrast, the partitioning shown in Figure \aref{fig:gnn-partitioning}{c} balances the workload and, at the same time, maximizes the localization of L-hop neighbors of training vertices, thereby reducing inter-partition communication.


Different from traditional data partitioning, GNN data partitioning has new goals. We summarize these goals as follows: 
\begin{itemize}[leftmargin=*]
\item \textbf{Goal 1: Minimize communication}. 
Partitioning label vertices and their L-hop neighbors into the same partition allows the L-hop sampled subgraphs to be distributed as locally as possible, thus avoiding extensive remote data requests.
\item \textbf{Goal 2: Balance computational load}.
Making labeled vertices and their L-hop neighbors evenly distributed across partitions allows a balanced computational load.
\item \textbf{Goal 3: Minimize total computational load}. 
Making the training L-hop subgraphs overlap with each other in a partition allows reusing the NN computation results, which helps reduce the total computational load.
\item \textbf{Goal 4: Balance communication load}. 
Communication load imbalance is a common challenge faced by both traditional graph computations and GNNs. However, compared to traditional graph computations, GNNs require frequent access to the L-hop neighbors of vertices, and the vertices of GNNs have high-dimensional feature representations. 
As a result, the problem of communication load imbalance is even more pronounced in GNNs. Making the remote neighbors of labeled vertices evenly across different partitions can balance the communication load.
\end{itemize}



\subsection{Existing Methods}
The graph partitioning methods used in the GNN systems can be categorized into the following three types:

\Paragraph{(1) Hash}
Hash is a general graph partitioning method that can satisfy various task requirements by defining different mapping rules, such as hashing by vertices \cite{Aligraph_VLDB19, Euler, AGL_VLDB20, ROC_MLSys20, DGCL_EuroSys21, Dorylus_OSDI21, NeutronStar_SIGMOD22, Legion_ATC23} or hashing by edges \cite{DistGNN_SC21, Sancus_VLDB22, Neugraph_ATC19, MariusGNN_EuroSys23}. 
P3 \cite{P3_OSDI21} employs a hash-based graph partitioning method that randomly assigns vertices to different partitions, achieving a balance in computational and communication loads (goal 2 and goal 4).
However, hash partitioning does not consider the labeled vertices in GNNs and their L-hop neighbors, so it cannot achieve the other two goals of GNN graph partitioning.

\Paragraph{(2) \Metisextended} 
Metis \cite{metis_siam98} partitions the graph with the goal of minimum edge cuts. In addition, it supports a variety of constraint mechanisms to extend Metis to meet different task requirements. We refer to this constrained Metis partitioning method as \Metisextended, which is used in DistDGL \cite{DistDGL_IAAA20}, and SALIENT++ \cite{SALIENTPlus_MLSYS23}.
The \Metisextended method effectively achieves the goals of GNN graph partitioning. 
Firstly, Metis employs a clustering algorithm for partitioning, ensuring that the neighbors of labeled vertices are allocated together, thus minimizing both computational and communication loads (goal 1 and goal 3). 
Secondly, Metis is extended by adding constraints with vertex masks to balance the number of labeled vertices (goal 2). In addition, Metis can be extended by adding constraints on vertex degrees to balance the number of edges across partitions, alleviating the load imbalance in computation and communication (goal 2 and goal 4).

\Paragraph{(3) Streaming}
Unlike traditional graph partitioning methods, streaming partitioning does not require storing the entire graph data but dynamically partitions the input vertices or edges.
Streaming partitioning can flexibly support various graph partitioning tasks by setting different score functions when assigning vertices or edges.
PaGraph \cite{PaGraph_SoCC20} and ByteGNN \cite{ByteGNN_VLDB22} both employ streaming graph partitioning methods. When assigning vertices, they prioritize assigning vertices to the partitions with the highest number of connected edges, aiming to minimize communication overhead (goal 1). Additionally, they use a factor to balance the number of label vertices (goal 2). 
However, these two streaming graph partitioning methods do not consider the density of partitioned graphs and the distribution of L-hop neighbors and thus suffer from high computational and imbalanced communication loads.

\newcolumntype{L}[1]{>{\raggedright\let\newline\\\arraybackslash\hspace{0pt}}m{#1}}
\newcolumntype{C}[1]{>{\centering\let\newline\\\arraybackslash\hspace{0pt}}m{#1}}
\newcolumntype{R}[1]{>{\raggedleft\let\newline\\\arraybackslash\hspace{0pt}}m{#1}}

\begin{table}[t]
\vspace{-0.22in}
\caption{Summary of evaluated partitioning methods.}
\label{table:partition-eval-system}
\vspace{-0.1in}
\footnotesize
\setlength{\tabcolsep}{1mm}{
\begin{tabular}{c|L{3cm}|c|c|c|c|c} 
    \hline
    \textbf{Method} & \multicolumn{1}{c|}{\textbf{Strategy}} & \textbf{\makecell[c]{Representative\\System}} & \textbf{G1} & \textbf{G2} & \textbf{G3} & \textbf{G4}\\
    \hline
    \multirow{1}{*}{Hash} 
    & Randomly assign vertices or edges.
    & \makecell[c]{P3  \cite{P3_OSDI21} } & \faBattery[0] & \faBattery[4] & \faBattery[0] & \faBattery[4]
    \\ \hline

    Metis-V 
    & Extend Metis by adding constraints on training vertex masks.
    & \makecell[c]{N/A}  & \faBattery[4]  & \faBattery[2] & \faBattery[2] & \faBattery[0]
    \\ \hline

    Metis-VE 
    & Extend Metis by adding constraints on training vertex masks and vertex degrees.
    & \makecell[c]{DistDGL \\ \cite{DistDGL_IAAA20}} & \faBattery[3] & \faBattery[3] & \faBattery[1] & \faBattery[1]
    \\ \hline
    
    Metis-VET 
    & Extend Metis by adding constraints on training/validation/test vertex masks and vertex degrees.
    & \makecell[c]{SALIENT \\++ \cite{SALIENTPlus_MLSYS23}} & \faBattery[3] & \faBattery[3] & \faBattery[1] & \faBattery[1]
    \\ \hline

    Stream-V
    & Assign vertex $v$ to a partition $P$ that has the most edges connected to $v$, while balancing the number of train vertices and caching L-hop neighbors.
    & \makecell[c]{PaGraph \\ \cite{PaGraph_SoCC20}} & \faBattery[4] & \faBattery[2] & \faBattery[0] & \faBattery[4]
    \\ \hline
    
    Stream-B  
    & Assign a block $B$ of vertices to a partition $P$ that has the most edges connected to $B$, while balancing the number of train/val/test vertices.
    & \makecell[c]{ByteGNN \\ \cite{ByteGNN_VLDB22}}  & \faBattery[2] & \faBattery[2] & \faBattery[0] & \faBattery[0]
    \\ \hline
    
\end{tabular}
}
\vspace{-0.2in}
\end{table}

\begin{figure*}[!t]
\vspace{-.15in}
\begin{minipage}[t]{.48\textwidth}
\centering
\subfloat[Aamzon]{\vspace{-0.13in}\label{fig:partition-comp-load-amazon}\includegraphics[width=1.18in]{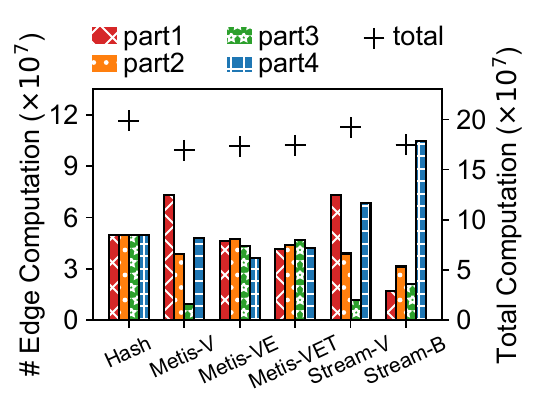}}
\subfloat[Products]{\vspace{-0.13in}\label{fig:partition-comp-load-products}\includegraphics[width=1.1in]{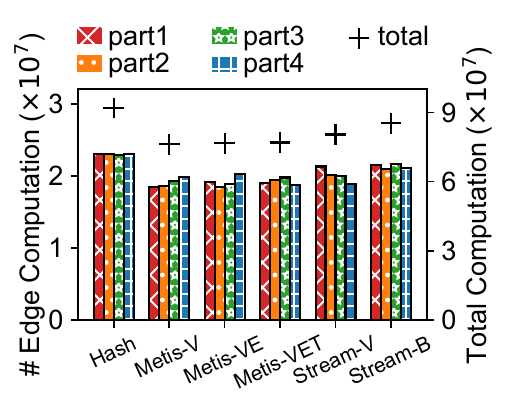}}
\subfloat[Reddit]{\vspace{-0.13in}\label{fig:partition-comp-load-reddit}\includegraphics[width=1.1in]{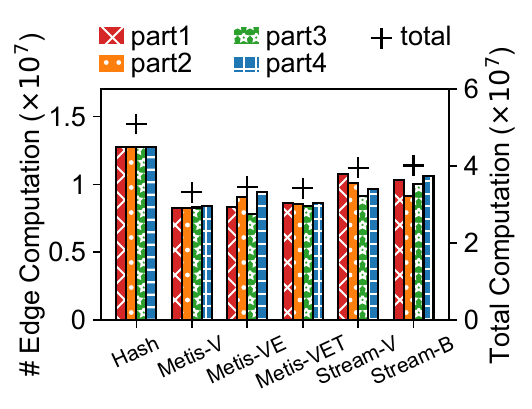}}
\vspace{-0.15in}
\caption{Computational load of different partitionings.}
\label{fig:partition-comp-load}
\end{minipage}
\hspace{0.05in}
\begin{minipage}[t]{.48\textwidth}
\centering
\subfloat[Aamzon]{\vspace{-0.13in}\label{fig:partition-comm-load-amazon}\includegraphics[width=1.1in]{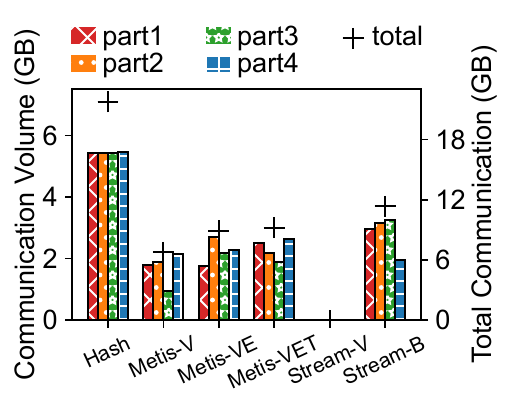}}
\subfloat[Products]{\vspace{-0.13in}\label{fig:partition-comm-load-products}\includegraphics[width=1.13in]{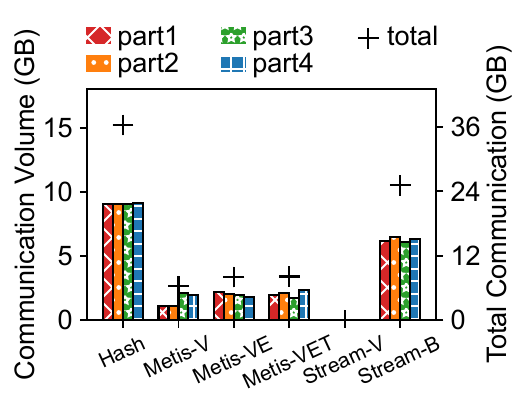}}
\subfloat[Reddit]{\vspace{-0.13in}\label{fig:partition-comm-load-reddit}\includegraphics[width=1.1in]{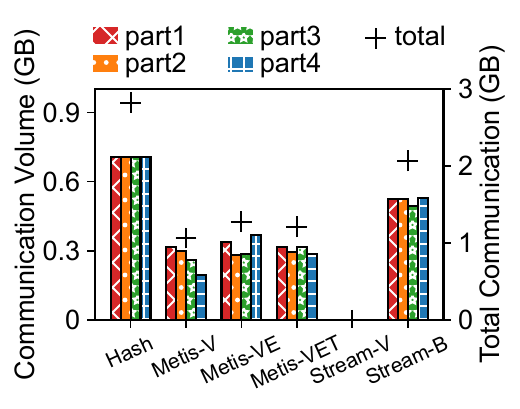}}
\vspace{-0.15in}
\caption{Communication load of different partitionings.}
\label{fig:partition-comm-load}
\end{minipage}
\vspace{-0.15in}
\end{figure*}

\subsection{Evaluation Results}

\Paragraph{Evaluated Methods} 
We conduct experimental analysis on the three types of data partitioning, i.e., Hash, \Metisextended, and Streaming. Table \ref{table:partition-eval-system} summarizes the partitioning methods and the systems that adopt these methods.
For the hash partitioning method, P3 \cite{P3_OSDI21} is a representative system that randomly partitions the vertices. 
For the \Metisextended partitioning method, DistDGL \cite{DistDGL_IAAA20} and SALIENT++ \cite{SALIENTPlus_MLSYS23} are two representative systems.
DistDGL balances the number of training vertices and edges in partitions, which we refer to as Metis-VE. 
SALIENT++ not only balances the training vertices and edges but also considers balancing the number of validation and test vertices, ensuring high performance in the inference stage, and we refer to this partitioning method as Metis-VET. 
In addition, we also compare the \Metisextended data partitioning method that only balances training vertices, which we refer to as Metis-V.
Although Metis-V is not employed by any existing GNN system for graph partitioning, we use Metis-V as a control group to comprehensively evaluate the performance of Metis-extend under different constraints.
PaGraph \cite{PaGraph_SoCC20} and ByteGNN \cite{ByteGNN_VLDB22} are two representative systems for the streaming partitioning method. PaGraph performs partitioning at the vertex level, while ByteGNN partitions at the block level. We refer to these two partitioning methods as Stream-V and Stream-B, respectively.

\subsubsection{Computational Workload Balance}
\label{sec-partition-comp}
We first conduct an experimental analysis from the perspective of computational workload balancing.
The computational workload comprises two parts: sampling and training. Regarding sampling, for remote-dependent vertices, sampling requests are sent to remote machines to be executed, and then the sampling results are returned. Therefore, the sampling workload on each machine includes both processing local sampling requests and processing remote sampling requests.
Regarding training, GNN training includes two parts: graph aggregation and NN computation. Since the graph aggregation dominates the overall computational cost. we only count the number of graph aggregations in the training part.

As illustrated in Figure \ref{fig:partition-comp-load}, regarding the hash partitioning method, since vertices are randomly assigned to different partitions with equal probability, it exhibits the most balanced computational workload. However, due to the neglect of vertex dependencies during partitioning, hash partitioning results in the highest total computational workload.
Regarding the \Metisextended partitioning, Metis-V only balances the number of training vertices, ignoring the distribution of L-hop neighbors of training vertices, resulting in load imbalance issues. By introducing vertex degree constraints to extend Metis, Metis-VE and Metis-VET alleviate the load imbalance problem observed in Metis-V. 
\Metisextended uses clustering for graph partitioning. When sampling vertices within clusters, different sampled vertices can share a large number of repeated neighbors \cite{2PGraph_TC23}. Hence, the higher the clustering degree in graph partitioning, the lower the computational load. Metis-VE and Metis-VET introduce additional constraints, which reduce the degree of clustering in the Metis partitioning. Therefore, the total computational load is higher than that of Metis-V.
Regarding the streaming partitioning, Stream-V and Stream-B aim to minimize cross-partition edges while balancing the number of labeled vertices when allocating vertices or blocks. This partitioning method performs well on non-power-law graphs but suffers from computational load imbalance on power-law graphs (i.e., Amazon) because it ignores the distribution of L-hop neighbors during partitioning.
To illustrate this point, we first compute the clustering coefficient of each partition graph to represent the density of the partitioned graphs. Then, we compute
the variance of the clustering coefficients of the partition graph to evaluate their distribution.
The results show that the variance of the clustering coefficient \cite{ClusterCoeffiecient_nature98} of the Hash partition graph is only $3.6\times 10^{-6}$, while the variances of Stream-V and Stream-B are 0.01 and 0.03, respectively. 
Therefore, the imbalance of partitioned graphs makes Stream-V and Stream-V suffer from computational load imbalance.


\subsubsection{Communication Workload Balance}
\label{sec-partition-comm}
We also conduct an experimental analysis of these graph partitioning methods from the perspective of communication workload balancing. The communication data comprises two parts: remote sample subgraphs and corresponding vertex features.
As shown in Figure \ref{fig:partition-comm-load}, regarding the hash partitioning, it exhibits the most balanced communication workloads, but the overall communication volume is also the highest. This is because hash partitioning ignores vertex dependencies during partitioning, resulting in a significant number of remote data requests and thus higher network communication overhead.
Regarding the \Metisextended partitioning, it can achieve the goal of minimizing communication (i.e., goal 1) by using a cluster-based partitioning method. 
However, it does not consider how to balance the communication load during the partitioning process and therefore suffers from an imbalanced communication load. 
Metis-V has the lowest total communication volume because its partitioned graph achieves the best clustering. 
By introducing vertex degree constraints to balance the number of edges in the partitioned graph, Metis-VE and Metis-VET alleviate the communication load imbalance problem but increase the total communication volume.
Regarding streaming partitioning, since Stream-V caches the graph data of L-hop neighbors of training vertices, no communication is required. 
Stream-B assigns a block to the partition with the most connections to it, which reduces the overall communication but ignores communication load balancing.


\begin{figure}[t]
\includegraphics[width=2.8in]{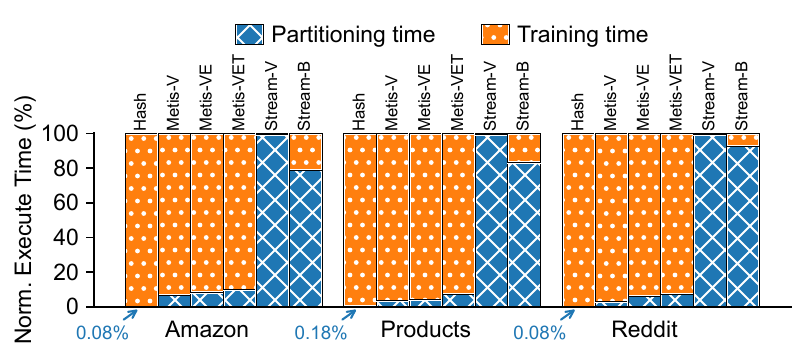}
\vspace{-0.15in}
\caption{Partitioning time vs. training time.}
\label{fig:partition-time-training-time}
\vspace{-0.17in}
\end{figure}

\begin{figure}[t]
\centering
\includegraphics[width=3.3in]{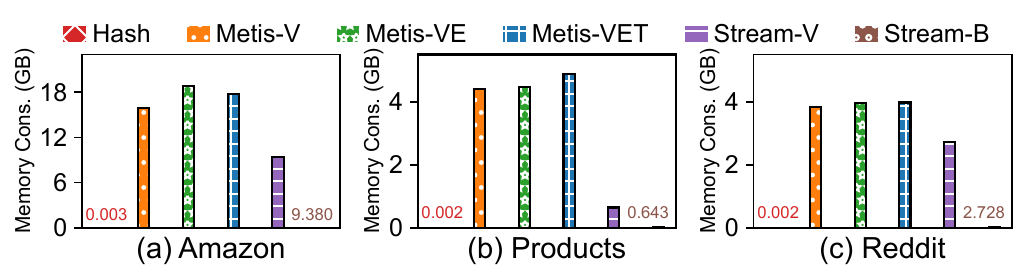}
\vspace{-0.15in}
\caption{Memory consumption of vary partitioning methods.}
\label{fig:partition-memory-consumption}
\vspace{-0.15in}
\end{figure}


\subsubsection{Partitioning Time vs. Training Time}
\label{sec-partiton-time-train-time}


\newcommand{\HashPartitionTime}{1.08s\xspace}
\newcommand{\ByteGNNPartitionTime}{4604.41s\xspace}

Although graph partitioning is a preprocessing task that only needs to be performed once before training, in practical applications, the time required cannot be ignored. 
In order to analyze the cost of graph partitioning, we show the proportion of graph partitioning time and training time on different datasets. As shown in Figure \ref{fig:partition-time-training-time}, 
regarding hash partitioning, it is the fastest and takes the smallest proportion of overall training time (only 0.11\% on average), resulting in the lowest partitioning overhead.
Regarding \Metisextended partitioning, the graph partitioning time portion for Metis-V, Metis-VE, and Metis-VET is 4.33\%, 6.11\%, and 7.98\%, respectively.
As more constraints are introduced, the partitioning overhead of \Metisextended gradually increases. Overall, the partitioning overhead of \Metisextended is deemed acceptable, constituting less than 10\% of the total training time.
Regarding streaming partitioning, it exhibits the longest graph partitioning time. The graph partitioning time portion for Stream-V and Stream-B is 99.36\% and 84.87\%, respectively.
This is because streaming-based partitioning involves extensive set intersection computations during vertex assignment, incurring substantial computational overhead. Additionally, due to vertex dependencies and the constraints of streaming graph partitioning, it cannot benefit from parallel acceleration.





\subsubsection{Partitioning Memory Consumption}
\label{sec-partiton-memory-usage}

Memory consumption is also a factor that should be considered when choosing a graph partitioning method. 
We conduct experiments to evaluate the memory consumption of partitioning methods. As shown in Figure~\ref{fig:partition-memory-consumption}, 
hash partitioning has the least memory consumption because it only needs to randomly assign vertices to different partitions and does not need to maintain the state of the partitions.
Streaming partitioning requires maintaining the state of the partition (e.g., allocated vertices) to guide the allocation of the next vertex or edge, so its memory consumption is higher than hash partitioning. In addition, the memory consumption of Stream-V is much higher than Stream-B. 
This is because Stream-V needs to obtain the L-hop neighbors of vertices when assigning vertices, which results in large memory consumption.
The Metis-extend partition has the highest memory consumption because it requires iteratively performing coarsening and uncoarsening steps, which results in severe memory overhead.

\subsubsection{Effect to Accuracy and Convergence Speed}
\label{sec-effect-to-accuracy}
In distributed training, each partition stores a portion of the graph data, and the training subgraphs are constructed only for local training vertices. Different graph partitioning methods result in various data distributions among partitions. For instance, in hash-based partitioning, vertices in each partition are randomly distributed, while in \Metisextended partitioning, vertices within partitions are distributed in a clustered manner. Hence, a question arises: \textit{does graph partitioning impact the accuracy and convergence speed of model training?}

\begin{figure}[t]
\centering
\includegraphics[width=3.3in]{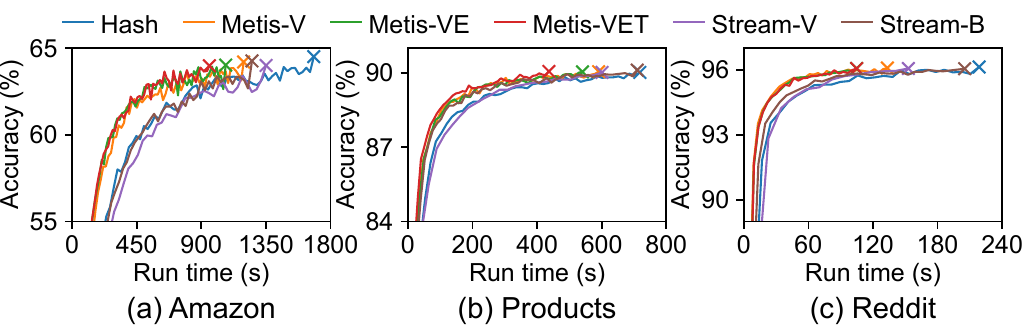}
\vspace{-0.12in}
\caption{Accuracy and convergence speed of vary partitioning methods.}
\label{fig:partition-convergence}
\vspace{-.15in}
\end{figure}

\begin{figure}[t]
\centering
\includegraphics[width=3.3in]{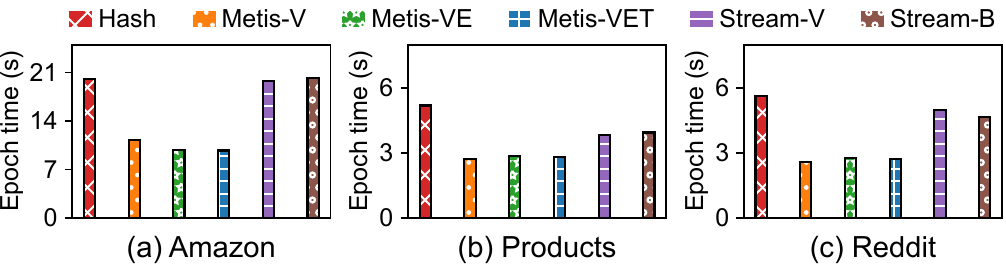}
\vspace{-0.12in}
\caption{Per-epoch runtime under different partition methods.}
\label{fig:partition-epoch-time}
\vspace{-0.15in}
\end{figure}

We compare the accuracy and convergence speed of different graph partitioning methods on three datasets. For each graph partitioning method, we train the model to converge and record the highest validation accuracy and run time.
Regarding model accuracy, as shown in Table \ref{table:partition-model-acc}, graph partitioning does not affect the highest accuracy.
Compared with DNN, there are complex dependencies between GNN data. Even when data is divided into multiple partitions, data exchange is still necessary between partitions due to the dependencies among vertices. Therefore, graph partitioning does not result in the loss of complete graph information and thus does not affect the final model accuracy.
Regarding convergence speed, as shown in Figure \ref{fig:partition-convergence}, hash partitioning exhibits the slowest convergence speed, followed by Stream-B. Figure \ref{fig:partition-epoch-time} further illustrates that Hash, Stream-V, and Stream-B have the longest per-epoch runtime, indicating that they require more time for the same number of iterations. 
Regarding the three \Metisextended graph partitioning methods, Metis-VET achieves the fastest convergence speed, followed by Metis-VE and Metis-V. Figure \ref{fig:partition-epoch-time} shows that the per-epoch runtime for each graph partitioning method is similar, but the convergence speed of Metis-VET is significantly faster than the other two methods.
We think that this is related to the characteristics of Metis. In mini-batch training, the training process needs to select a batch of vertices from local training vertices. Since Metis employs a clustering method to partition densely connected vertices together, this reduces the randomness in batch vertex selection, which in turn affects the model's convergence speed. 
Since Metis-VET has the most constraints, it has the least clustering effect, resulting in the fastest convergence speed.



\begin{table}[t]
\caption{Model accuracy under different partition methods.}
\label{table:partition-model-acc}
\vspace{-0.1in}
\footnotesize
\setlength{\tabcolsep}{.7mm}{
\begin{tabular}{c|c|c|c|c|c|c|c} 
    \hline
    & \textbf{Hash} & \textbf{Metis-V} & \textbf{Metis-VE} & \textbf{Metis-VET} & \textbf{Stream-V} & \textbf{Stream-B} & \textbf{Diff.}\\
    \hline
    Reddit & 96.2\% & 96.2\% & 96.2\% & 96.1\% & 96.4\% & 96.2\% & $\pm 0.3\%$\\ 
    \hline
    Products & 90.1\% & 90.3\% & 90.2\% & 90.2\% & 90.5\% & 90.2\% & $\pm 0.4\%$\\ 
    \hline
    Amazon & 64.5\% & 64.2\% & 64.4\% & 64.7\% & 65.1\% & 64.3\% & $\pm 0.9\%$\\
    \hline
\end{tabular}
}
\vspace{-0.15in}
\end{table}


\subsection{Lessons Learned}

\begin{enumerate}[leftmargin=*]
\item Existing graph partitioning methods fail to meet the graph partitioning requirements for GNN training, which poses new challenges different from traditional graph computation, such as minimizing total computational load and balancing communication load. (\S\ref{sec-partiton-goal-challenge})

\item There is a trade-off between minimizing computational load and achieving load balance when partitioning a graph. Partitioning densely connected vertices together can effectively reduce computation but may disrupt load balancing. (\S\ref{sec-partition-comp})

\item Communication load imbalance is more serious in GNN training than in traditional graph computation due to the high-dimensional vertex features and embeddings.
Most existing GNN systems ignore this issue. (\S\ref{sec-partition-comm})

\item Most graph partitioning overhead is acceptable for GNN training. Although streaming partitioning is more flexible, existing methods \cite{PaGraph_SoCC20, ByteGNN_VLDB22} suffer from high computational costs and inefficient implementation due to low parallelism. (\S\ref{sec-partiton-time-train-time})

\item 
Although Metis-extend has better performance, its graph partitioning requires a large memory consumption, especially for large-scale graphs. This limitation prevents the scalability of the Metis-extend partitioning method. (\S\ref{sec-partiton-memory-usage})

\item Shorter per-epoch runtime does not necessarily mean faster convergence speed. Partitioning densely connected vertices into the same partition can significantly reduce inter-partition communication and decrease the per-epoch runtime but may decrease the randomness of model training, thus affecting convergence speed. (\S\ref{sec-effect-to-accuracy})


\end{enumerate}

\vspace{-.05in}
\section{Batch Preparation}
\label{sec_data_sampling}
\subsection{Goals and Challenges}

After the graph partitioning step, sample-based mini-batch training need a batch preparation step before training starts, which determines the model quality and training performance.
In the batch preparation step, the training process divides the training vertices into multiple batches and then samples the $L$-hop subgraphs for the training vertices based on a sampling method.
The batch size determines the number of batches (i.e., the frequency of model parameter updates) in an epoch. The sampling method determines graph topology information that GNN models can learn.

However, there is a trade-off between accuracy and performance when choosing batch size and sampling method.
\begin{enumerate}[leftmargin=*]
\item Batch size. Due to vertex dependencies, vertices within a batch usually have a large number of common neighbors. As the batch size increases, the number of these common neighbors also grows. 
The sampling and NN computation results of these common neighbors can be reused, thus reducing the computational load of an epoch.
However, increasing the batch size reduces the frequency of model parameter updates, which may affect accuracy and overall convergence speed.

\item Sampling method. The sampling method of GNN aims to reduce the size of the training graph to support large-scale GNN training under limited memory. The training speed of GNN can be significantly improved by reducing the size of the sampled subgraph. However, as the size of the sampled subgraph decreases, the information learned by the model each time will also decrease, impacting the accuracy and convergence rate.
\end{enumerate}

\noindent This section empirically studies the impact of batch size and sample method on accuracy and performance.

\begin{table}[t]
\caption{Default settings of batch size and sampling parameters in existing GNN systems.}
\vspace{-0.1in}
\label{table:bs-sr-set}
\footnotesize
\setlength{\tabcolsep}{3.9mm}{
\begin{tabular}{c|c|c|c} 
    \hline

    System & Batch size & Fanout & Sampling rate \\ \hline 
    P3 \cite{P3_OSDI21} & 1000 & (25, 10) & N/A \\ \hline 
    DistDGL \cite{DistDGL_IAAA20} & 2000 & \makecell{(25, 10) \\ (15, 10, 5) } & N/A \\ \hline 
    PaGraph \cite{PaGraph_SoCC20} & 6000 & (2, 2) & N/A \\ \hline 
    GNNLab \cite{GNNLab_EuroSys22} & 8000 & \makecell{(10, 25) \\ (15, 10, 5)} & N/A \\ \hline 
    ByteGNN \cite{ByteGNN_VLDB22} & 512 & (10, 5, 3) & N/A \\ \hline 
    BNS-GCN \cite{BNS-GCN-MLSYS22} & full & N/A & 0.1 \\ \hline 
    SALIENT++ \cite{SALIENTPlus_MLSYS23} & 1024 & \makecell{(25, 15)\\ (15, 10, 5)} & N/A \\ \hline 

\end{tabular}}
\vspace{-0.2in}
\end{table}

\subsection{Existing Methods}
\Paragraph{Batch Size} 
GNN training methods can be categorized into two types: full-batch training and mini-batch training. 
Full-batch training requires all vertices to participate in training simultaneously, it requires more storage and computing resources. In addition, the model parameters are updated only once within an epoch, which results in slower model convergence. 
In contrast, mini-batch training involves only a subset of vertices in each training iteration, allowing model parameters to be updated multiple times in an epoch.

\Paragraph{Sampling Fanout and Sampling Rate} The sampling methods of GNN can be divided into two types: fanout-based \cite{DGL_Arxiv19, PaGraph_SoCC20, P3_OSDI21, Sancus_VLDB22} and ratio-based \cite{Aligraph_VLDB19, DistDGL_IAAA20, BNS-GCN-MLSYS22}. Fanout-based methods generate sampled subgraphs based on a fixed number, while ratio-based methods generate sampled subgraphs based on ratio. These two sampling methods determine the size of the sampled subgraph and are orthogonal to sampling algorithms (e.g., vertex-wise sampling \cite{GraphSage_NIPS17, PinSage_KDD18, VRGCN_ICML20}, layer-wise sampling \cite{FastGCN_ICLR18, ASGCN_NIPS18, LADIES_nips19}, and subgraph-wise sampling \cite{ClusterGCN_KDD19, GraphSAINT_ICLR20}).

To the best of our knowledge, the settings of these parameters have not yet been fully discussed. Users typically configure batch size and sampling methods based on their experience and specific task requirements. Table \ref{table:bs-sr-set} summarizes parameter settings for several systems. Regarding batch size, common choices include 512, 1024, 2000, 6000, and 8000. Regarding fanout, in 2-layer GNN models, it is typically set to (25, 10), while for 3-layer models, (15, 10, 5) is commonly chosen.
Regarding sampling rate, BNS-GCN \cite{BNS-GCN-MLSYS22} recommends using a 0.1 ratio to sample boundary vertices (i.e., vertices involved in cross-partition communication).
In this section, we conduct extensive experiments to analyze the training performance of the GNN under different parameter settings and also propose a hybrid training method to accelerate the convergence of the model.

\Paragraph{Other Optimizations}
We note that some work to optimize batch preparation from a system optimization perspective.
For example, shared-memory parallel batch preparation \cite{SALIENT_MLSYS22} uses zero-copy communication to access the remote feature across training processes.
However, these optimizations are orthogonal to our concerns, so we do not evaluate them experimentally.

\begin{figure}[!t]
\vspace{-.15in}
\centering
\subfloat[Reddit]{\vspace{-0.05in}\includegraphics[width=1.3in]{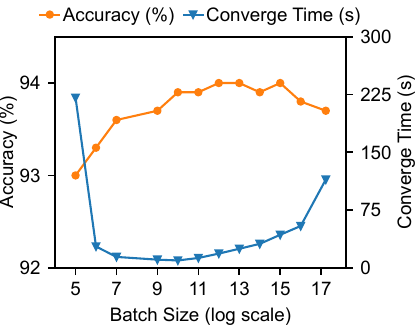}} 
\hspace{0.2in}
\subfloat[Products]{\vspace{-0.05in}\includegraphics[width=1.3in]{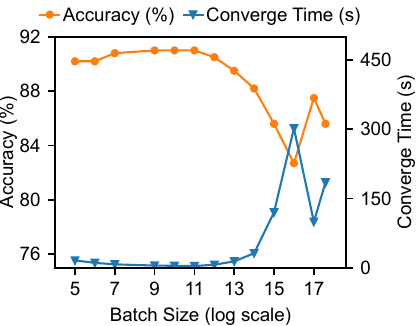}}
\vspace{-0.15in}
\caption{Accuracy and convergence speed with varying batch sizes.}
\label{fig:acc-speed-nts-adam}
\vspace{-0.15in}
\end{figure}

\begin{figure}[!t]
\begin{minipage}[t]{.48\textwidth}
\centering
\subfloat[Reddit]{\vspace{-0.05in}\includegraphics[width=1.3in]{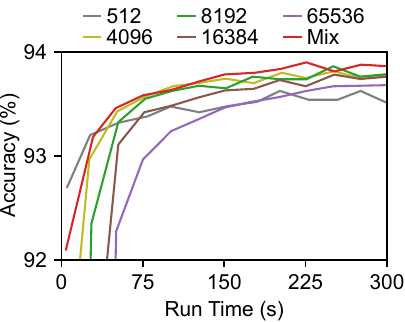}}\hspace{0.1in}
\hspace{0.2in}
\subfloat[Products]{\vspace{-0.05in}\includegraphics[width=1.3in]{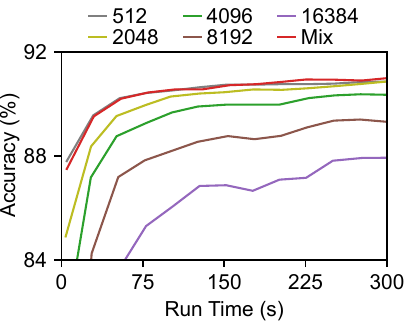}}\hspace{1pt}
\vspace{-0.15in}
\caption{Performance with adaptive batch size.}
\label{fig:mix-bs-size}
\end{minipage}
\vspace{-0.2in}
\end{figure}

\subsection{Evaluation Results} 
\subsubsection{Batch Size}
\label{sec:batch-size-analysis}



To investigate the impact of batch size on training, we compare the training accuracy and convergence speed of GNN models under different batch sizes. As shown in Figure \ref{fig:acc-speed-nts-adam}, we find that different batch sizes have a significant impact on both the accuracy and convergence speed. The specific phenomenon can be summarized in two points:
\begin{enumerate}[leftmargin=*]
\item Reducing the batch size will speed up convergence. However, after exceeding a lower batch size, the convergence speed of the model decreases.

\item Increasing the batch size increases the accuracy of the model. However, after exceeding a higher batch size, the model's accuracy decreases.
\end{enumerate}

\noindent Taking the Reddit dataset as an example, 
regarding convergence speed, the model converges faster as the batch size decreases. 
For example, when the batch size is reduced from 32,768 to 128, the model's convergence speed increases by 3.02 times. 
However, if the batch size continues to decrease, e.g., from 128 to 64, the model converges 1.95 times slower.
Regarding accuracy, the accuracy of the model increases with batch size.
For example, when the batch size increases from 32 to 4096, the accuracy of the model improves by 1\%.
However, if the batch size continues to increase, such as from 4096 to 196615, the model's accuracy decreases by 0.3\%.

In addition, we find that small batch sizes have larger gradient magnitudes, and large batch sizes have smaller gradient magnitudes. Therefore, we try to explain these two phenomena by the training characteristics of Mini-Batch Gradient Descent \cite{RESNET_MGD_CVPR16} (MGD). 
For phenomenon (1), we think that a larger gradient magnitude is good for finding the direction of the model's optimal point faster, but too large a gradient magnitude is unfavorable for model convergence.
For phenomenon (2), we think that a smaller gradient magnitude is good for finding the optimal point of the model, but too small gradient magnitudes tend to fall into the local optimal point of the model.
To ensure both accuracy and convergence speed, we can first use a large gradient magnitude (small batch size) to find the optimal point direction and then use a small gradient magnitude (large batch size) to close the optimal point.

\Paragraph{Adaptive Batch Size}
\label{sec:adaptive-batch-size}
Previous work used fixed batch sizes for GNN training. Through experiments, we find that smaller batch sizes have faster convergence and larger batch sizes have higher accuracy. Based on this, we propose an adaptive batch size training method, which dynamically combines the advantages of different batch sizes during training.
The strategy for switching batch sizes is to use a small batch size at the beginning of training so that the model can quickly converge to a higher accuracy. Then, the batch size is gradually increased to converge to the highest accuracy.
For example, on the Reddit dataset, we first start training with a batch size of 512 (because it has the fastest convergence rate) and then gradually increase the batch size until the batch size reaches 8,192.
As illustrated in Figure \ref{fig:mix-bs-size}, adopting this adaptive training method significantly accelerates the model's convergence rate. On the Reddit and Products datasets, the convergence speed increased by \mixBatchSizeReddit and \mixBatchSizeProducts times, respectively. 




\begin{figure}[!t]
\vspace{-.15in}
\centering
\subfloat[Reddit]{\vspace{-.05in}\includegraphics[width=1.3in]{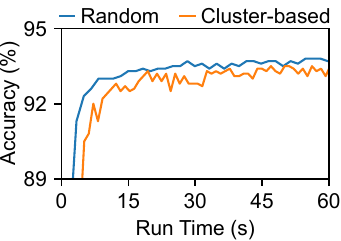}} 
\hspace{0.1in}
\subfloat[Products]{\vspace{-.05in}\includegraphics[width=1.3in]{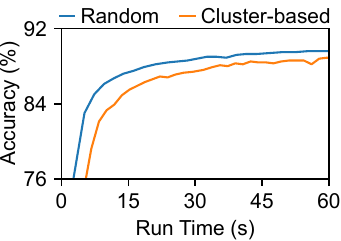}}
\vspace{-.15in}
\caption{Accuracy and convergence speed of random and cluster-based batch selections.}
 \label{fig:batch-type-acc}
\end{figure}

\subsubsection{Batch Selection}
\label{sec:batch-selection}
Batch selection determines the vertices that each batch contains in training. 
Selecting batch vertices with bias increases the variance in model training, thereby affecting the model's convergence.
Furthermore, the density of connections between vertices within the same batch can impact the overall computational load.
To investigate the influence of different batch selection methods on accuracy and convergence speed, we compare two existing batch selection methods.
1) \textit{Random} \cite{DistDGL_IAAA20, PyG_ICLR19, SALIENTPlus_MLSYS23, PaGraph_SoCC20, GNNLab_EuroSys22}. It randomly selects a batch of vertices. 
2) \textit{Cluster-based} \cite{2PGraph_TC23}. It leverages graph clustering algorithm (e.g., Metis \cite{metis_siam98}) to arrange batches.

\begin{table}[!t]
\vspace{-0.05in}
\caption{Per-epoch runtime of different batch selection methods.}
\label{table:batch-select-epochtime-VE}
\vspace{-0.15in}
\footnotesize
\begin{tabular}{c|c|c|c|c} 
    \hline
    Dataset & Method & Per-epoch runtime (s) & Involved \#V & Involved \#E \\ \hline

    \multirow{2}{*}{Products}
    & random  & 2.4	& 24.8 M & 45.9 M \\ \cline{2-5}
    & cluster-based  & 1.0 & 15.7 M & 38.3 M \\ \hline

    \multirow{2}{*}{Reddit} 
    & random  & 1.7 & 6.8 M & 25.2 M \\ \cline{2-5} 
    & cluster-based  & 0.6 & 4.2 M & 16.7 M \\ \hline

\end{tabular}
\vspace{-0.15in}
\end{table}

Table \ref{table:batch-select-epochtime-VE} shows the average per-epoch runtime and computational load for each batch selection method, with computational load measured by the number of vertices and edges involved in the training subgraph.
The cluster-based method results in the shortest per-epoch runtime. 
This method selects a batch of vertices from the cluster graph. The vertices in the cluster graph have denser connections, which increases the chance of reusing computational results.
The random batch selection method does not consider the dependencies between vertices when selecting batched vertices, it involves the highest number of vertices and edges in computations.

Then, we compare the accuracy of these two batch selection methods on the two datasets of Reddit and Ogbn-products.
As shown in Figure \ref{fig:batch-type-acc}, the random selection method has the highest accuracy because it has no bias when selecting samples.
Since cluster-based batch selection is limited to the internal structure of the clustered graph, it introduces data bias, resulting in lower accuracy. In addition, we observe that the training process of cluster-based is unstable.
We think this phenomenon is related to the distribution of batched subgraphs. 
Since each cluster graph has a different density, this may lead to an imbalanced distribution of the subgraphs involved in training.
To illustrate this point, we calculate the clustering coefficient for each batched subgraph as a measure of its density. Subsequently, we calculate the variance of the clustering coefficients for the batched subgraphs within an epoch to assess their distribution.
The experimental results show that the variance of the cluster-based method (${2\times10^{-4}}$) is much larger than that of the random selection method (${1.1\times10^{-6}}$).
Therefore, the imbalanced distribution of batch subgraphs leads to the training process being unstable.


\subsubsection{Fanout} 
\label{sec:fanout-analysis}

\begin{figure}[!t]
\vspace{-0.2in}
\centering
\subfloat[Fanout]{\vspace{-0.06in}\label{fig:fanout-exp}\includegraphics[width=1.3in]{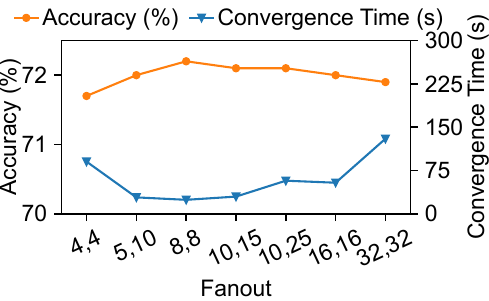}} 
\hspace{0.15in}
\subfloat[Sample Rate]{\vspace{-0.06in}\label{fig:sample-rate-exp}\includegraphics[width=1.3in]{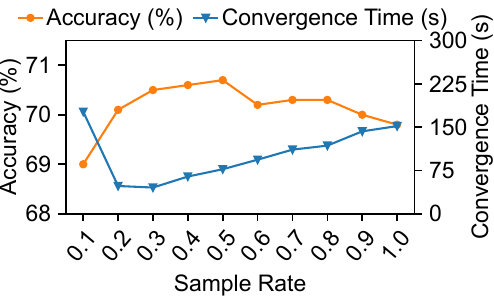}} 
\vspace{-0.15in}
\caption{Accuracy and convergence speed of different fanout settings and sample rate settings (Arxiv).}
\vspace{-0.1in}
\end{figure}

The fanout is another key parameter in GNN training.
To investigate the impact of fanout on training, we compare the accuracy and convergence speed of GNN models under different fanouts on the Ogbn-arxiv dataset.
As shown in Figure \ref{fig:fanout-exp}, the accuracy of the GNN model shows a trend of "first increase and then decrease" as the fanout increases, while the convergence speed shows the opposite.
We think this phenomenon is related to the degree of vertices. Most real-world graphs are skewed (e.g., social networks, e-commerce networks, and citation networks), which means that only a few vertices have very large degrees, while most vertices have small degrees. Hence, a fixed fanout is not conducive to these power-law graphs. We find that low-degree vertices are better suited for smaller fanouts, while high-degree vertices are more appropriate for larger fanouts.

To illustrate this point, we select four typical fanout based on the results in Figure \ref{fig:fanout-exp}: the smallest fanout (4, 4), the best-performing fanout (8, 8), and two larger fanout (16, 16) and (32, 32).
Then, we measure their prediction accuracy on low-degree and high-degree vertices using the Ogbn-arxiv dataset.
As shown in Table \ref{table:predict-acc-degree}, as fanout increases, the prediction accuracy on low-degree vertices decreases while the accuracy on high-degree vertices increases.
This is because, for low-degree vertices, using smaller fanouts can enhance the randomness of sampling while preserving the integrity of graph information. However, using smaller fanouts for high-degree vertices may result in too few neighbor samples, making it challenging to learn the complete graph structure during training.

\begin{table}[t!]
\caption{Accuracy of high and low degree vertices (Arxiv).}

\vspace{-0.1in}
\footnotesize
\label{table:predict-acc-degree}
\begin{tabular}{c|c|c|c|c} 
    \hline
    \multirow{2}{*}{\textbf{Vertex type}} & \multicolumn{4}{c}{\textbf{Fanout}} \\ 
    \cline{2-5}
    & 4, 4 & 8, 8 & 16, 16 & 32, 32\\ 
    \hline
    Low-degree vertices & 0.718 & \textbf{0.722} & 0.720 & 0.718 \\ 
    \hline 
    High-degree vertices & 0.786 & 0.794 & 0.810 & \textbf{0.817} \\ 
    \hline
\end{tabular}
\vspace{-0.1in}
\end{table}

\subsubsection{Sampling Rate} 
\label{sec-sample-rate}
Through the above experimental analysis on fanout, it is not optimal to use a fixed fanout for vertices of different degrees. 
In addition, we find that fanout is not flexible enough to express sampling, especially when sampling high-degree vertices. 
Sampling rate \cite{BNS-GCN-MLSYS22, ClusterGCN_KDD19, GraphSAINT_ICLR20} seems to be a good choice, and sampling vertices of different degrees proportionally is more fair than fanout.
As shown in Figure \ref{fig:sample-rate-exp}, we compare the training accuracy and convergence speed under different sampling rates. 
Similar to the experimental results of fanout, as the sampling rate increases, the accuracy shows a trend of "first increase and then decrease". 
It should be noted that the overall accuracy of the sampling rate is lower than that of fanout.
There are two main reasons, firstly, smaller sampling rates disadvantage low-degree vertices. For example, for a vertex with degree 20, if the sampling rate is set to 0.1, only 2 vertices can be sampled each time, which is far less than the number of fanout samples (10, 25). 
Secondly, increasing the sampling rate will sacrifice the randomness of sampling, thus affecting the accuracy and convergence speed of the model. Therefore, using only a sampling rate is not optimal.

\begin{table}[t!]
\vspace{-0.2in}
\caption{Accuracy and performance comparison of fanout-based sampling and fanout-rate hybrid sampling (Arxiv).}
\vspace{-0.1in}
\footnotesize
\label{tab:mix-fanout-samplerate}
\begin{tabular}{c|c|c|c|c|c|c} 
    \hline
    & 4, 4 & 8, 8 & 10,15 & 10,25 & 32, 32 & hybrid\\ 
    \hline
    Accuracy (\%) & 71.5 & 72.1 & 72.1 & 72.1& 71.5 & 72.1 \\ 
    \hline 
    Time (s) & 300 & 172 & 165 & 237 & 300 & \textbf{99} \\ 
    \hline
\end{tabular}
\vspace{-0.15in}
\end{table}


\Paragraph{Fanout-Rate Hybrid Sampling}
\label{sec:fanout-rate-hybrid}
Previous work used fixed fanout or sampling rate for GNN training.
Through experiments, we found that low-degree vertices are better suited for smaller fanouts, while high-degree vertices are more appropriate for larger fanouts. In addition, as the sampling rate increases, the model accuracy shows a trend of "first increase and then decrease".
Based on this, we propose a hybrid training method that combines fanout and sampling rate.
The fundamental strategy of the hybrid training method is to employ fanout sampling for low-degree vertices and sampling rate sampling for high-degree vertices.
As shown in Table \ref{tab:mix-fanout-samplerate}, using a hybrid training method can significantly speed up the convergence of the model. Compared to using only fanout (8, 8), the convergence speed is improved by \mixFanoutSamplerateSpeedup times.

\subsection{Lessons Learned}

\begin{enumerate}[leftmargin=*]

\item There is a trade-off between accuracy and performance when setting a batch size. By adaptively setting a batch size (using a smaller size at the beginning and then gradually increasing it), it can effectively accelerate the convergence. (\S\ref{sec:batch-size-analysis}) 

\item Although the complex cluster-based batch selection method can reduce the per-epoch runtime, simple batch selection methods (e.g., random selection) perform better, because no bias is introduced when selecting data samples, resulting in higher accuracy and faster convergence. (\S\ref{sec:batch-selection})

\item It is nonsense to perform sampling with a fixed fanout on all vertices. It is suggested to set a smaller fanout for low-degree vertices and a larger fanout for high-degree vertices. (\S\ref{sec:fanout-analysis})

\item Dynamically adjusting the batch size during training and combining fanout and sampling rate during sampling can significantly improve the convergence speed without sacrificing accuracy. (\S\ref{sec:adaptive-batch-size} and \S\ref{sec:fanout-rate-hybrid})

\end{enumerate}

\section{Data Transferring}
\label{sec_data_movement}

\subsection{Goals and Challenges}




After the batch preparation step
, we need to extract the features of the selected vertices in a batch and then load the batch data to the GPU for NN computation.
The data transferring of GNN has the following special challenges.
\begin{enumerate}[leftmargin=*]
    \item \textbf{Irregular memory access}.
    Compared with DNN, the random memory access pattern of GNN causes a large amount of random I/O overhead in feature extraction.
    
    \item \textbf{Redundant communication}. There are a large number of repeated vertices and edges between sampled subgraphs due to vertex dependencies.
    Unlike graph computation, vertices and edges in GNNs are represented as feature vectors, which usually have thousands of dimensions. Transferring these large and redundant vertex/edge features increases the burden of data transferring. 
\end{enumerate}

\begin{figure}[t]
\vspace{-0.2in}
\includegraphics[width=3in]{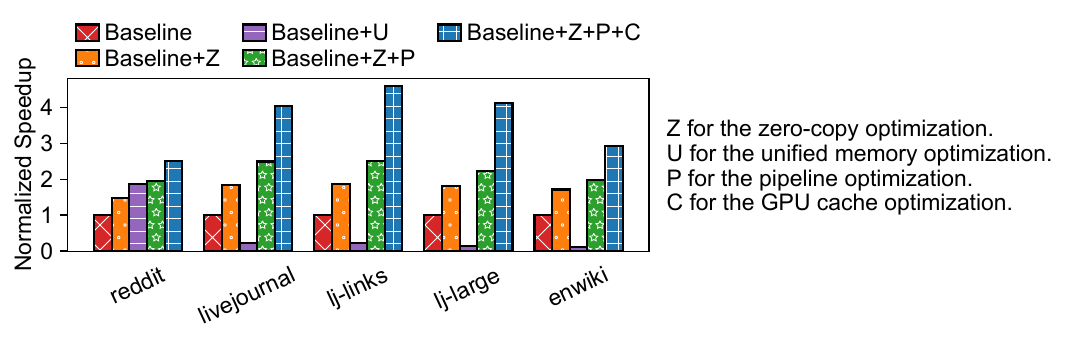}
\vspace{-0.17in}
\caption{Performance gain of data transfer optimizations.}
\label{fig:diff-optim}
\vspace{-0.2in}
\end{figure}

\subsection{Existing Methods}

The three following data transfer optimization methods have been widely employed in GNN systems \cite{DistDGL_IAAA20, PaGraph_SoCC20, Dorylus_OSDI21, GNNLab_EuroSys22, Sancus_VLDB22, SALIENTPlus_MLSYS23, Legion_ATC23, BGL_NIDS23}, including UVA-based data direct access, task pipelining, and cache-based data reusing.

\Paragraph{Data transferring method} 
The data transfer methods between CPU and GPU can be categorized as explicit and implicit transfer. 
In \textit{explicit} transfers, developers need to explicitly call the CUDA \cite{CUDA_doc} API to transfer data between the CPU and GPU memory. Explicit transfer is suitable for transferring large blocks of data, which can fully utilize the bandwidth. Therefore, for GNN training, scattered vertex features need to be extracted into a contiguous memory space and then transferred to the GPU.
In contrast, zero-copy \cite{zero_copy_doc} is an \textit{implicit} transfer method based on Unified Virtual Addressing (UVA), which allows GPU threads to access CPU memory directly, thus avoiding the overhead of extracting vertex features.
However, kernel performance suffers from high latency accesses to zero-copy memory over the PCIe. Therefore, it requires fine-grained data access orchestration to mitigate the performance impact of latency access \cite{pytorch_directed_VLDB21}.
Unified memory \cite{unified_memory_blog} is another \textit{implicit} memory access method that provides a single memory address space for GPU and CPU. The requested data memory pages are automatically migrated and cached to the device, and subsequent accesses to the same pages will read the data from GPU memory without additional data transfer. When the memory consumed by cached pages exceeds the GPU memory, some pages may need to be evicted from the GPU to make space for new pages.
However, due to the high cost of page migrations and the irregular access pattern of graphs, unified memory is not an efficient way of handling graph algorithms \cite{HyTGraph_ICDE23}.

\Paragraph{Task Pipelining}
Pipelining decomposes a task into multiple stages and allows these stages to be executed in parallel on different processors or resources. 
In a CPU and GPU heterogeneous training scenario, a complete batch training process consists of three steps: batch preparation, data transferring, and NN computation. These three steps are executed on three different devices, i.e., CPU, PCIe, and GPU. Due to the data dependency between these steps, the traditional training process performs these three steps sequentially \cite{DGCL_EuroSys21, PaGraph_SoCC20, P3_OSDI21}.
Since there is no obvious dependency between training in different batches, we can use pipeline optimization to improve hardware utilization. 
Specifically, when the CPU finishes sampling for batch $b$, it transfers the batch data to the GPU via PCIe, while the CPU starts sampling for batch $b+1$. Once the GPU receives the data for batch $b$, it begins NN computation. In this way, all hardware resources can be scheduled simultaneously, maximizing the use of hardware resources.

\Paragraph{Cache-based Data Reusing}
The training process needs to sample labeled vertices to generate L-hop training subgraphs. 
Due to the intricate dependencies between vertices, there are a large number of duplicate vertices between batch training subgraphs \cite{PaGraph_SoCC20}. This results in redundant data transfers between the CPU and GPU, which seriously wastes bandwidth resources.
GPU caching is a cache-based data reuse method \cite{PaGraph_SoCC20, Legion_ATC23, SALIENT_MLSYS22, BGL_NIDS23, Sancus_VLDB22}. Unlike zero-copy and pipeline optimization, by caching vertex features in GPU memory, the data transfer volume between the CPU and GPU can be fundamentally reduced.


\newcommand{\datatransferOverhead}{73.4\%\xspace}
\newcommand{\featureextractOverhead}{31.2\%\xspace}
\newcommand{\dataloadingOverhead}{42.2\%\xspace}

\subsection{Evaluation Results}
\subsubsection{Data transferring method}
\label{sec-uva}

As shown in Figure \ref{fig_exp_performance_breakdown}, the data transferring step dominates the entire training process (taking up \datatransferOverhead in total), with feature extraction accounting for \featureextractOverhead and data loading accounting for \dataloadingOverhead.
As shown in Figure \ref{fig:diff-optim}, compared to explicit data transfer (e.g., Baseline), zero-copy optimization (e.g., Baseline+Z) has \zeroSpeedup times the performance gain on average.
Using unified memory can achieve an improvement of 1.88 times on a small-scale dataset (Reddit). However, on large-scale datasets, using unified memory does not bring positive performance gains due to frequent page migration overhead.

\begin{figure}
\vspace{-0.2in}
\centering
\includegraphics[width=3in]{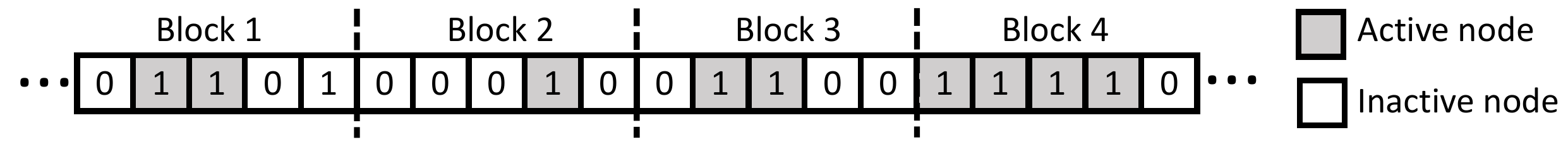}
\vspace{-0.15in}
\caption{An example of active vertices distribution in a batch.}
\label{fig:active-block}
\vspace{-0.25in}
\end{figure}


\noindent\textbf{Does Hybrid Data Transferring Help?}
Regarding data transfer optimization, a hybrid data transfer optimization method \cite{HyTGraph_ICDE23} was recently proposed. This approach combines explicit and implicit transferring methods, dynamically selecting the most suitable transfer method for each data block. 
In order to explore whether hybrid data transfer optimization can accelerate GNN training, we analyze the distribution of active vertices (sampled vertices) within each batch. As shown in Figure \ref{fig:active-block}, we count the active vertices in the data block in units of 256KB \cite{pytorch_directed_VLDB21}.
For each data block, we set a threshold. If the number of active vertices exceeds this threshold, the block will be considered suitable for explicit transfer. Otherwise, it is suitable for implicit transfer (e.g., when the threshold is 0.5, only block 4 is suitable for explicit transfer).

\begin{figure}[t]
\vspace{-.15in}
\centering
\includegraphics[width=1.15in]{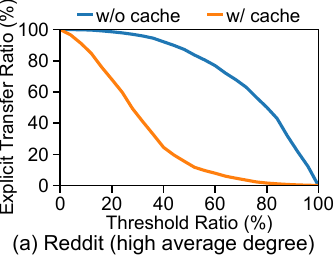}
\hspace{.2in}
\includegraphics[width=1.2in]{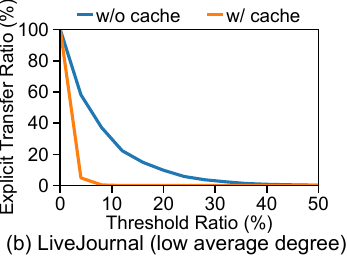}
\vspace{-0.12in}
\caption{Ratio of active blocks with different thresholds.}
\label{fig:hybrid-trans}
\vspace{-0.15in}
\end{figure}

Figure \ref{fig:hybrid-trans} shows the ratio of data blocks suitable for explicit transfer at different thresholds. We can observe that as the threshold increases, the ratio of data blocks suitable for explicit transfer decreases significantly.
Since the Reddit dataset has a high average degree, which makes the distribution of active vertices denser, it is more suitable for the explicit transfer of data blocks. 
However, after applying GPU cache optimization, the number of data blocks suitable for explicit transfer is significantly reduced (orange line in Figure \ref{fig:hybrid-trans}). For example, when the threshold is set to 80\%, only 2\% of the data blocks are suitable for explicit transfer even for the Reddit dataset.
Based on the above analysis, We find that the demand for hybrid data transferring in GNN training is not significant. This is due to the unique computational patterns of GNNs. 
Traditional graph computations have different vertex access patterns in iterative computations \cite{Gemimi_OSDI16, HyTGraph_ICDE23}. However, in GNN training, the distribution of vertices in the sampled subgraph is often random and fragmented, especially after implementing a caching optimization. 
Therefore, though the hybrid data transfer shows promising results in graph computation, it does not help GNN training.


\begin{figure}[t]
\includegraphics[width=3.1in]{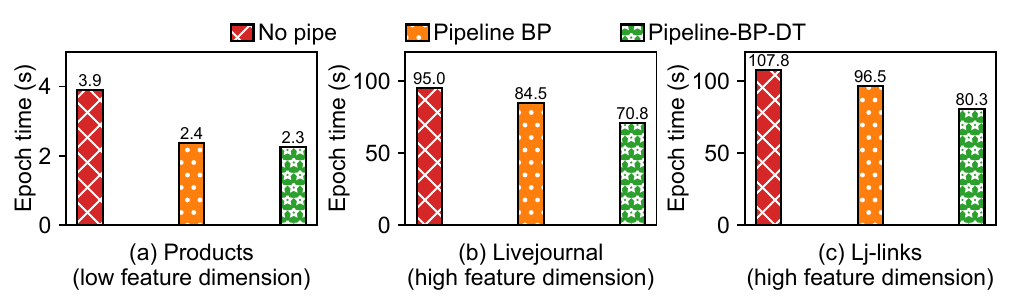}
\vspace{-0.15in}
\caption{Pipeline training ablation study.}
\label{fig:pipe-ablation}
\vspace{-0.25in}
\end{figure}

\subsubsection{Task Pipelining}
\label{sec-task-pipeline}
In the CPU-GPU heterogeneous training scenarios, a complete batch training process can be divided into three steps: batch preparation (BP), data transfer (DT), and NN computation (NN).
As shown in Figure \ref{fig:diff-optim}, Baseline+Z+P represents the simultaneous use of zero-copy transfer optimization and pipelining optimization. It offers a performance improvement of \ZPSpeedup times compared to Baseline (i.e., explicit transfer). When compared to Baseline+Z (i.e., zero-copy optimization), the pipeline optimization brings an additional performance gain of \pipelineSpeedup times.

\eat{
To provide a more detailed analysis of the pipelining performance, we perform pipelined ablation experiments. Figure \ref{fig:pipe-ablation} reports the per-epoch runtime when adding each step to the pipeline.
}
Figure \ref{fig:pipe-ablation} is a pipeline ablation experiment that reports the per-epoch runtime when adding each step to the pipeline.
No Pipe is the slowest one due to its totally sequential execution.
Pipeline BP can significantly reduce the running time by pipelining the BP step.
The results show that the acceleration effect of pipelining is not significant because data transfer is still the main bottleneck. On the Livejournal and Lj-links datasets, the time proportion of data transfer is 58.8\% and 53.1\%, respectively, which dominate the training process. 
Some research works \cite{LazyGCN_NeurIPS20, G3_SIGMOD_2023, MGG_OSDI_2023} have demonstrated that improving the performance of the pipeline is possible by balancing GNN workloads between CPUs and GPUs and employing fine-grained task scheduling.

\subsubsection{Cache-based Data Reusing}
\label{sec-cache}
\begin{figure}[t]
\vspace{-.15in}
\centering
\includegraphics[width=1.2in]{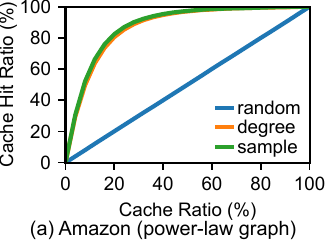}
\hspace{.2in}
\includegraphics[width=1.2in]{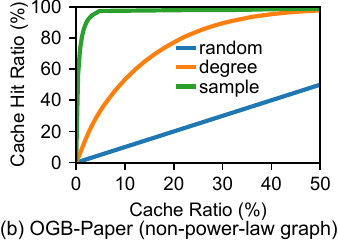}
\vspace{-0.15in}
\caption{Performance comparison of caching policies.}
\label{fig:diff-cache}
\vspace{-0.2in}
\end{figure}

There are two main caching strategies: 
(1) degree-based caching (degree), 
(2) pre-sampling-based caching (sample).
The degree-based caching strategy prioritizes caching vertices with large out-degrees. The sampling-based caching strategy first obtains the access frequency of each vertex by pre-sampling and then prioritizes the caching of frequently accessed vertex features.
We compare the performance of these two caching strategies.
As shown in Figure \ref{fig:diff-cache}, 
on the Amazon dataset, the performance of these two caching strategies is comparable. 
However, on the OGB-Paper dataset, the pre-sampling-based caching strategy significantly outperforms the degree-based caching strategy.
This is because degree-based caching is based on the assumption that higher-degree vertices are more likely to be sampled. Therefore, degree-based caching strategies perform well on power-law graphs but perform poorly on non-power-law graphs. In addition, the degree-based caching strategy is only applicable to the uniform vertex sampling algorithm \cite{GraphSage_NIPS17}. For special sampling algorithms (such as importance sampling \cite{FastGCN_ICLR18}), the degree-based assumption is no longer valid. In contrast, the pre-sampling-based caching strategy has better robustness and performs well on both power-law graphs and non-power-law graphs.


\vspace{-.1in}
\subsection{Lessons Learned}

\begin{enumerate}[leftmargin=*]

\item Due to GNN's irregular data access pattern, the vertex features required for heterogeneous training are often scattered in memory. Through UVA-based data direct access (zero-copy) optimization, the GPU device can access CPU memory directly, avoiding expensive feature extraction operations. (\S\ref{sec-uva})

\item Vertices do not exhibit an obvious active or deactivated state in GNN training, the hybrid data transfer optimization will not help improve data transferring. (\S\ref{sec-uva})

\item The effect of pipeline optimization is not outstanding (less than 50\% improvement in most cases) under our experimental settings because the data transferring step is the main bottleneck, overwhelming the other tasks. (\S\ref{sec-task-pipeline})

\item GPU caching is the most significant data transfer optimization method as it fundamentally reduces data transfer volume.
Compared to degree-based caching, pre-sampling-based caching exhibits better robustness. (\S\ref{sec-cache})

\end{enumerate}

\vspace{-.1in}
\section{conclusion}
\label{sec_conclusion}
In this paper, we conduct a comprehensive evaluation of the optimization techniques adopted by existing GNN systems from a data management perspective.
We show that graph partitioning of GNN is a complex task. Achieving the goals of GNN graph partitioning presents significant challenges. 
In addition, we find that there are significant differences in the accuracy and convergence speed of GNN under different parameter settings. We propose adaptive batch size training and hybrid fanout and sample rate training methods to accelerate convergence while maintaining accuracy. 
Lastly, we conduct evaluations of three data transfer optimizations, summarizing their advantages, disadvantages, and suitable scenarios.
Based on these results, we provide insights and lessons learned for enhancing the training performance of GNNs.



\vspace{-0.10in}
\begin{acks}
This work is supported by the National Key R\&D Program of China (2018YFB1003400), 
the National Natural Science Foundation of China (U2241212, 62072082, 62202088, 62072083, and 62372097), and the Fundamental Research Funds for the Central Universities (N2216015 and N2216012).
Yanfeng Zhang and Qiange Wang are the corresponding authors.
\end{acks}

\balance

\bibliographystyle{ACM-Reference-Format}
\bibliography{ref}

\end{document}